\documentclass[final]{ws-procs11x85}
\usepackage{ws-procs-thm}           

\usepackage[main=english]{babel} 
\usepackage{microtype} 
\usepackage[babel,autostyle]{csquotes} 
\usepackage{booktabs,multirow}

\newcommand{\EE}{\ensuremath{\mathbb{E}}}
\newcommand{\PP}{\ensuremath{\mathbb{P}}}
\newcommand{\RR}{\ensuremath{\mathbb{R}}}

\providecommand{\norm}[1]{\ensuremath{\lVert#1\rVert}}

\newif\ifarxiv
\arxivtrue

\begin{document}

\title{Quantifying surprise in clinical care: Detecting highly informative events in electronic health records with foundation models}

\author{Michael C. Burkhart, Bashar Ramadan, Luke Solo, William F. Parker, and Brett K. Beaulieu-Jones}
\address{Department of Medicine, University of Chicago,\\ Chicago, Illinois, USA \\
    E-mail: \{burkh4rt,basharramadan,lsolo,wparker,beaulieujones\}@uchicago.edu
}

\begin{abstract}
We present a foundation model-derived method to identify highly informative tokens and events in electronic health records. Our approach considers incoming data in the entire context of a patient's hospitalization and so can flag anomalous events that rule-based approaches would consider within a normal range. We demonstrate that the events our model flags are significant for predicting downstream patient outcomes and that a fraction of events identified as carrying little information can safely be dropped. Additionally, we show how informativeness can help interpret the predictions of prognostic models trained on foundation model-derived representations.
\end{abstract}




\ifarxiv
	\copyrightinfo{\copyright\,\the\year{} The Authors. Distributed under the terms of the Creative Commons Attribution Non-Commercial (CC BY-NC) 4.0 License.}
\else
	\clearpage
\fi

\section{Introduction}

Healthcare generates a stream of data, including vitals, labs, medications, and respiratory support. Clinical decision making requires parsing and understanding this information and its importance in the context of each patient's medical history. Oftentimes, event summaries like automatically-collected vitals provide little additional knowledge about a patient~\cite{Xu25,And23}. Clinicians are commonly notified regardless, resulting in increased cognitive burden and alarm fatigue~\cite{JC13}. For over a decade now, the Joint Commission has included ``reduc[ing] patient harm associated with clinical alarm systems'' as a National Patient Safety Goal~\cite[NPSG.06.01.01, since 2014]{JC24}. In this paper, we investigate the extent to which foundation model (FM)-derived estimates of event information can be used to highlight the most important events in a patient's record. In essence, we explore which events are surprising to the FM based on a comparison between what the model expects to happen next and what is observed. Identifying important or surprising events has the potential to substantially improve our understanding of healthcare delivery and better inform clinicians about the status of their patients. 

Divergence between the model's expectation and the actual observation or informativeness broadly indicates one of three things: (1) practice variation, when a clinician makes a decision which deviates from what is typically done in similar contexts in the training data (e.g., prescribing a medication off-label); (2) an unexpected change in patient condition that would generally be observed in clinical measurements (e.g., a lab result which indicates a change in patient state that could not be predicted using observed covariates); or (3) issues of data quality which could be present in either orders or patient measurements (e.g., a typo when entering a value). In all three cases, there is the potential to learn substantially from the model's ``surprise,'' to potentially reduce clinical errors, and to rapidly and succinctly surface the most important information for a clinician. This could be used to both better inform the time-sensitive decision-making that often occurs in the hospital setting as well as to provide a summary of the most important information about a patient for downstream analyses like phenotyping or sub-population identification.

While this work focuses on patients receiving critical care in the inpatient setting, the approach aims to be one which can be generalized to many healthcare settings with longitudinal data and outcomes. This work operates at the level of Electronic Health Records (EHRs) corresponding to individual hospitalization stays. For each hospitalization, we form a sequence of tokens that describe the admitted patient, along with their admission type, and then chronicle vitals, administered medications, lab results, assessments, and a few other categories of data as they become available, ending with a token for discharge~\cite{Ste21,McD23,Bur25}. We perform self-supervised training of a foundation model (FM) to predict the next token in one of these sequences given all previous tokens. Such models have proven remarkably effective across a number of  fields~\cite{Awa25,Raz25,Sun25} but most importantly in our case for predicting a variety of downstream clinical outcomes~\cite{Wor23b,Wor23}. Furthermore, these models are generative~\cite{Ng02} in the sense that they estimate the joint probability distribution on these sequences. Given a trained model and a novel sequence, we can estimate the context-aware (conditional) information of each token. We call a series of tokens that become available at the same time ``an event'' and calculate context-aware information for each event. We show that highly informative tokens and events are more predictive of downstream outcomes and tend to result in greater changes to the model-derived understanding of a patient's current condition.

In this paper, we present the first comprehensive study of FM-derived information quantification for tokens and events in EHRs. Our main contributions are as follows:

\begin{enumerate}
	\item We propose a principled FM-derived method to identify highly informative tokens and events in a patient's EHR. As opposed to classical rules-based methods, our context-informed approach identifies anomalous labs and assessments even when a patient has values within what would be considered a normal range. As opposed to the variable importance methods applied to classifiers trained for specific outcomes, our method defines informativeness in terms of the sequences themselves.
	\item We illustrate how the occurrence of highly informative events impacts a patient's prognosis and alters the FM-derived representation that is commonly used for making downstream predictions. In terms of interpretability, this allows us to provide a list of events deemed most informative to the FM. We show that dropping these events from a patient's timeline impacts the performance of downstream prognostic models. Conversely, we show that events carrying the least information can be dropped without sacrificing predictive performance.
\end{enumerate}

\section{Related work}

Early approaches to modeling sequential data derived from EHRs focused on recurrent neural networks (RNNs) including Long Short-Term Memory~\cite[LSTM]{Hoc97} networks~\cite{Lip17,Cho16,Bea18,Raj18}. Approaches shifted from RNNs to transformers~\cite{Vas17} beginning with variations on BERT~\cite{Dev19}, including BEHRT~\cite{Li20} and Med-BERT~\cite{Ras21}. Subsequently, Foresight~\cite{Kra24} and ETHOS~\cite{Ren24,Ren25} both used generative pretrained transformer~\cite[GPT]{Rad18} architectures. Wornow, et al. provided a detailed review of FMs for EHRs up to 2023~\cite{Wor23b}. More recently, Mamba~\cite{Gu24}, a selective state-space model, has found applications in ClinicalMamba~\cite{Yan24} and EHRMamba~\cite{Fal24}.

Some efforts have been made to better understand these types of models. Beaulieu-Jones, et al.~\cite{Bea21} noted that sequential EHR models can learn both from the patient's actual state (e.g. the result of a particular lab) and from clinicians' actions (e.g. the fact that a particular lab was ordered). They found that models trained on demographics, admissions data, and charges from the first day of admission (clinician-initiated actions) performed competitively against models trained on full sequences of EHR data. In doing so, they raised an important point about understanding which tokens and events in a patient's sequence drive a model's understanding of that sequence.

Wornow et. al~\cite{Wor25} studied, among other things, how sequences derived from EHRs differ from natural language (written English). They showed how EHRs exhibit copy-forwarding of chronic diagnoses, irregular spacing between tokens, and increased perplexity of tokens over time due to disease progression. Their definition of perplexity relates closely to our definition of informativeness, but they did not investigate which types of tokens tend to carry more information, nor did they consider subsequences. In contrast to this work, they focused on much longer-term time horizons consisting of multiple clinical encounters, whereas we focus on single hospitalization events.

\section{Methods}

\subsection{Data}

We considered 422,765 hospitalization events for adults (age 18 or older) from the Beth Israel Deaconess Medical Center between 2008–2019 (MIMIC-IV-3.1~\cite{Joh23}) and 50,440 hospitalization events from the UCMC health system between March 2020 and March 2022. We restricted our analysis to patients with stays of at least 24 hours. We formatted EHR data from each health system into the CLIF standard~\cite{Roj25}. The MIMIC patients were partitioned into training, validation, and test sets at a 70\%-10\%-20\% rate. We then collected each hospitalization event for patients in a given set. In this way, hospitalization records in the test set corresponded to patients with no hospitalization events in the training or validation sets to avoid any potential information leakage. The UCMC data was primarily used as a held-out test set. For this reason, we partitioned the internal patients into training, validation, and test sets at a 5\%-5\%-90\% rate according to the time of their first hospitalization event, with training patients coming first, followed by validation, and then test.

\subsection{Tokenization}

\begin{figure}[htb]
	\centering
	\includegraphics[width=\textwidth]{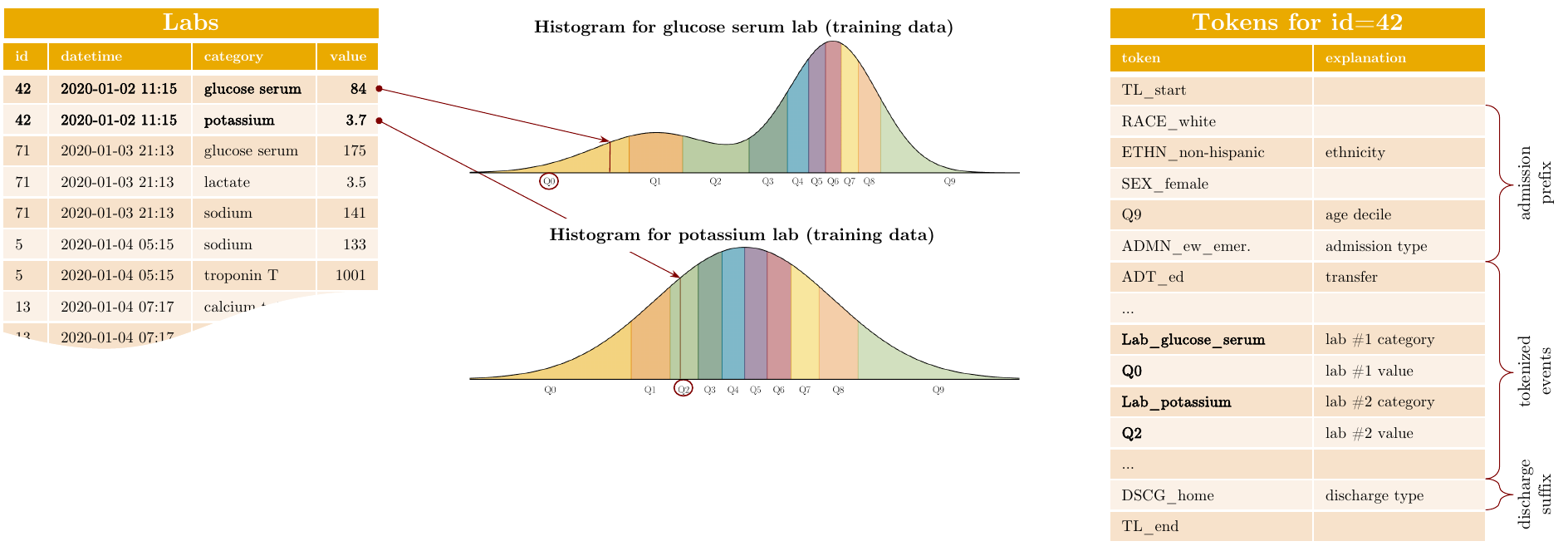}
	\caption{\emph{Category-value tokenization.} We convert lab results into tokens as follows. For each lab category, we determine decile cutoffs (center) using all results corresponding to that lab category available in the training dataset. Each lab value is then encoded as a decile (with \texttt{Q0} corresponding to the lowest decile, \texttt{Q1} to the next, and so on up to \texttt{Q9}) and inserted into the corresponding hospitalization in temporal order.}
	\label{fig:catval-tokenization}
\end{figure}

We converted each hospitalization event from the CLIF standard into a sequence of tokens (represented computationally as non-negative integers) as follows. For a given sequence, the first token always corresponds to a timeline start token. The next three tokens contain patient-level demographic information on race, ethnicity, and sex. The following two tokens correspond to admission-specific information, namely patient age converted to a decile and admission type. Taken together, we refer to the 5 tokens occurring immediately after the timelines start token as the \emph{admission prefix}. Tokens corresponding to a variety of events for a hospitalization are then inserted in the same order in which these events occurred. Transfers are encoded with their standardized location category. Labs are encoded with two tokens and inserted at the time results become available: one for the lab category, and a second corresponding to the deciled lab value in the training data within that category. We call this strategy, of tokenizing categories and binning their corresponding values according to the training value of the deciles, category-value tokenization.  See Figure~\ref{fig:catval-tokenization} for an illustration. A handful of other tables receive this type of tokenization: vitals and results according to vital category, medication and dosage by medication category, assessment and results by assessment category. Respiratory information is recorded at the beginning of respiratory support; the encoded information is mode category and device category. We include a token indicating if a patient is placed into a prone position. All hospitalization-related data is encoded this way and inserted in chronological order. Tokens that arrive synchronously correspond to an event and always appear coterminously in a sequence. Timelines then end with a token for discharge category and a dedicated timeline end token. We did not use time-spacing or artificial time tokens~\cite{Pan21} as recent studies suggest they do not improve performance~\cite{Wor25}.

\subsection{Context-aware information}
Consider the set $V^T$ of length-$T$ sequences of tokens drawn from some vocabulary $V$. Such sequences correspond directly to tokenized EHR data as described in the previous subsection. For a given sequence $x=(x_1,\dotsc, x_T)$ and indices $1\leq u \leq v \leq T$, we let $x_{u:v}=(x_u, x_{u+1}, \dotsc, x_v)$ correspond to the subsequence and $x_{<u}=x_{1:u-1}$ to the context at $u$ for $u>1$. If $p$ is a probability distribution on $V^T$, we let $p(x_{u:v})=\PP_{X\sim p}(X_{u:v}=x_{u:v})$ denote the marginal distribution and $p(x_{u:v}|x_{y:z})=\PP_{X\sim p}(X_{u:v}=x_{u:v}| X_{y:z}=x_{y:z})$ denote the conditional for indices $u,v,y,z$. We adopt the convention that $p(x_{u:v} | x_{<1}) = p(x_{u:v})$. With these definitions, the Shannon self-information~\cite{Sha48} of a certain realized subsequence $x_{u:v}$ under $p$ is given by $I_{p}(x_{u:v}) = - \log_2 p(x_{u:v}).$ The \emph{context-aware information} associated to a realized subsequence $x_{u:v} \in V$ and context $x_{<u} \in V^{u-1}$ is defined analogously, by
\begin{equation}
	\label{eq:cond_inf}
	\boxed{I_{p}(x_{u:v} | x_{<u}) = - \log_2 p(x_{u:v} | x_{<u}).}
\end{equation}
In the case of a single token $x_t$, we have $t=u=v$ and refer to
\begin{equation}
	\label{eq:tokenwise_cond_inf}
	\boxed{I_{p}(x_t | x_{<t}) = -\log_2 p(x_t | x_{<t})}
\end{equation}
as tokenwise context-aware information. As $p(x_{u:v} | x_{<t}) = \prod\nolimits_{t=u}^v p(x_t | x_{<t})$, it follows that
\begin{equation}
	\label{eq:inf_add}
	I_{p}(x_{u:v} | x_{<u}) = \textstyle\sum\nolimits_{t=u}^v I_{p}(x_t | x_{<t}).
\end{equation}
Thus, context-aware information is additive.

This quantity plays a pivotal role in the training of standard models. A model is a parameterized distribution $p_\theta$ on $V^T$. Training attempts to find parameters $\theta$ that minimize the relative entropy (or Kullback–Leibler divergence~\cite{Kul51}) between  the empirical distribution $\hat p$ given by the training set and $p_\theta$,
\begin{equation}
	\label{eq:kl}
	D( \hat p || p_\theta )
	= \underbrace{\EE_{X_{1:T}\sim \hat p} [I_{p_\theta}(X_{1:T})]}_{H(\hat p, p_\theta)} - \underbrace{\EE_{X_{1:T}\sim \hat p} [ I_{\hat p}(X_{1:T})]}_{H(\hat p)}.
\end{equation}
Here, $H(\hat p, p_\theta)$ is the cross-entropy between $\hat p$ and $p_\theta$ and $H(\hat p)$ is the entropy of $\hat p$. As this latter term is independent of $\theta$, it may be disregarded during training / optimization. By~\eqref{eq:inf_add}, we have the simplification
$H(\hat p, p_\theta)
	= \textstyle\sum\nolimits_{t=1}^T \EE_{X_{1:T}\sim \hat p} [ I_{p_\theta}(X_t | X_{<t})]$.
We see that the training process optimizes $\theta$ to minimize the expected tokenwise context-aware information over the training set. In more general terms, training finds the model $p_\theta$ that makes the training set least surprising. This is equivalent to maximum likelihood estimation~\cite[\S5.5]{Goo16}. Upon completion of training, $p_{\theta_*}$ with optimized parameters $\theta_*$ serves as our best approximation to $p$ and can be used to calculate the context-aware information \eqref{eq:cond_inf} in new timelines for tokens and subsequences.

\subsection{Model}
For our parameterized distribution $p_\theta$ on sequences of tokens/integers, we train a model from scratch based on the Llama-3.2 model architecture~\cite{Gra24} with a hidden size of 1024, intermediate size of 2048, 8 hidden layers, and 8 attention heads, for a total of 67.3 million parameters. Wornow, et al.'s~\cite[Fig. 1B]{Wor25} architecture comparison indicates that the Llama architecture performs favorably to GPT~\cite{Rad18}, Hyena~\cite{Pol23}, and Mamba~\cite{Gu24} architectures for context lengths of 1000-2000 tokens, such as we use here.

\subsection{Training}
As our vocabulary is created during the tokenization process, we train models from scratch, as opposed to fine-tuning models that have been pre-trained on a tokenized natural language vocabulary. We train weights to minimize~\eqref{eq:kl} with AdamW~\cite{Los19}, a variant of Adam~\cite{Kin15} with decoupled weight decay~\cite{Han88}. Training batches were formed by packing tokenized sequences into a $b\times 1024$-dimensional array in row-major order where $b$ is the batch size.\footnote{Note, because of this packing strategy, the model does not learn that our sequences always start with the timeline start token. By convention, the true $p(x_1)$ is an indicator function on the timeline start token, so that $I_p(x_1)$ should be $-\log_2 1 = 0$. (Deterministic tokens do not carry information.) The model should learn that the first token after the start token should be a race token, and then an ethnicity token, and so on, because for these predictions, context is supplied.} We used tree-structured Parzen estimators~\cite{Aki19} to tune the learning rate (between $5\cdot 10^{-5}$ and $5\cdot 10^{-4}$, inclusive) and effective batch size (between 32 and 96, inclusive). Models were trained on a single compute node with 8×A100 (40GB PCIe) GPUs, connected with 2×16-core 3.0-GHz AMD Milan processors. The model having best-performing loss on the MIMIC evaluation set was selected and provides the $p$ used to calculate context-aware information for the remainder of the paper.

\subsection{Representation-based prognostic models}
As a causal language model or state space-based model processes a sequence $x_{1:T}$ of tokens, it forms a representation $R(x_{1:t})\in \RR^d$ of the subsequence encountered up to the $t$th token for each $1\leq t \leq T$, where $d$ tends to be at least a few hundred dimensions. In Llama models, we take the last hidden state to be our representation, with $d$ equal to the ``hidden size'' parameter, in our case set to 1024. In many FM-based works\cite{Ste24,Fal24,Wor25}, these representations or a function of them provide the basis for all subsequent prognostic predictions. For example, the representation $R(x_{1:t_0})$ of a patient's timeline that contains all tokens occurring prior to some cutoff time will then be used as features in a logistic regression model to predict outcomes for that patient occurring after the cutoff time.  All patients start with the same representation, i.e. $R(x_1)$ corresponds to the representation associated to the ``timeline start'' token. As tokens are added to each timeline, these representations diverge. We are generally interested in the relationship between informativeness and corresponding changes in representation space at both the token and event levels of granularity. Establishing a strong relationship between information content and changes in representation could help to explain the predictions of representation-based prognostic models. To this end, we define the magnitude of the change in representation space when token $x_t$ is added as
\begin{equation}
	\label{eq:delta_t}
	\Delta_t=\norm{R(x_{1:t})-R(x_{1:t-1})}
\end{equation}
where the norm is taken to be the standard Euclidean norm and define the path length in representation space corresponding to a subsequence $x_{u:v}$ as
\begin{equation}
	\label{eq:delta_uv}
	\Delta_{u:v} = \textstyle\sum\nolimits_{t=u}^v \Delta_t.
\end{equation}

\subsection{Highlighting examples}
We denote the $i$th timeline in the test set by $x^{(i)}=x^{(i)}_{1:T}$. For each $1\leq t \leq T$, we use the model to calculate the associated tokenwise context aware information, $I_p(x^{(i)}_t|x^{(i)}_{<t})$. For events, i.e. maximal contemporaneous subsequences $x_{u:v}$ occurring between the prefix and suffix tokens, we calculate eventwise context-aware information $I_p(x^{(i)}_{u:v}|x^{(i)}_{<u})$ as in~\eqref{eq:cond_inf}.

\subsection{Redaction experiment}
\label{ss:redaction}

We restrict our cohort to patients who are admitted to the ICU within the first 24 hours of their admission. We consider two outcomes: \emph{inpatient mortality}, defined as patient death prior to discharge from the hospital, and \emph{long length-of-stay}, defined as discharge occurring $\ge7$ days after admission.

For each timeline truncated at the 24 hour mark, we calculate context-aware information for each event. For each of 10\%, 20\%, 30\%, \& 40\%, we drop that percentage of either the most or the least informative events, or that percentage of events chosen at random. We do this for each combination of percentage and method (most, least, random), creating 12 partially redacted versions of the original 24-hour timelines.

For each data version, we extract 24-hour representations $R(x_{1:t_0})$ using our model, where $t_0\leq 1024$ corresponds to the last token to arrive within 24 hours of admission. Note that, in an abuse of notation, $t_0$ depends on the hospitalization sequence $x$. Much more information is collected for some patients in the first 24 hours than for others. We then train a logistic regression classifier to predict each outcome (inpatient mortality and long length-of-stay) given the 24-hour representations on the training portion of the MIMIC dataset. We apply each model to the respective versions of both the MIMIC and UCMC test sets.

We perform bootstrap resampling to estimate 95\% confidence intervals for test set-based variability in the ROC-AUC~\cite[\emph{cf.} \S13.3]{Efr93}. This method takes a fixed classification model and forms an empirical distribution of performance metrics by resampling test data 10,000 times.

We also use bootstrap sampling to estimate $p$-values for the hypothesis test of $H_0: \text{AUC}_0=\text{AUC}_1$ against the one-sided alternative $H_a: \text{AUC}_0 >\text{AUC}_1$, where $\text{AUC}_0$ corresponds to the original AUC and $\text{AUC}_1$ to the AUC from a fixed classifier built and tested on redacted timelines~\cite[Algorithm 16.1]{Efr93}. This method compares the observed difference in AUC performance against differences obtained from 10,000 resamplings under the hypothesis that predictions from the two classifiers are exchangeable. This bootstrapping approach also only simulates variability in the test set given fixed classifiers.

\section{Results}

\subsection{Highlighted timelines}
\begin{figure}[htb]
	\centering
	\includegraphics[width=\textwidth]{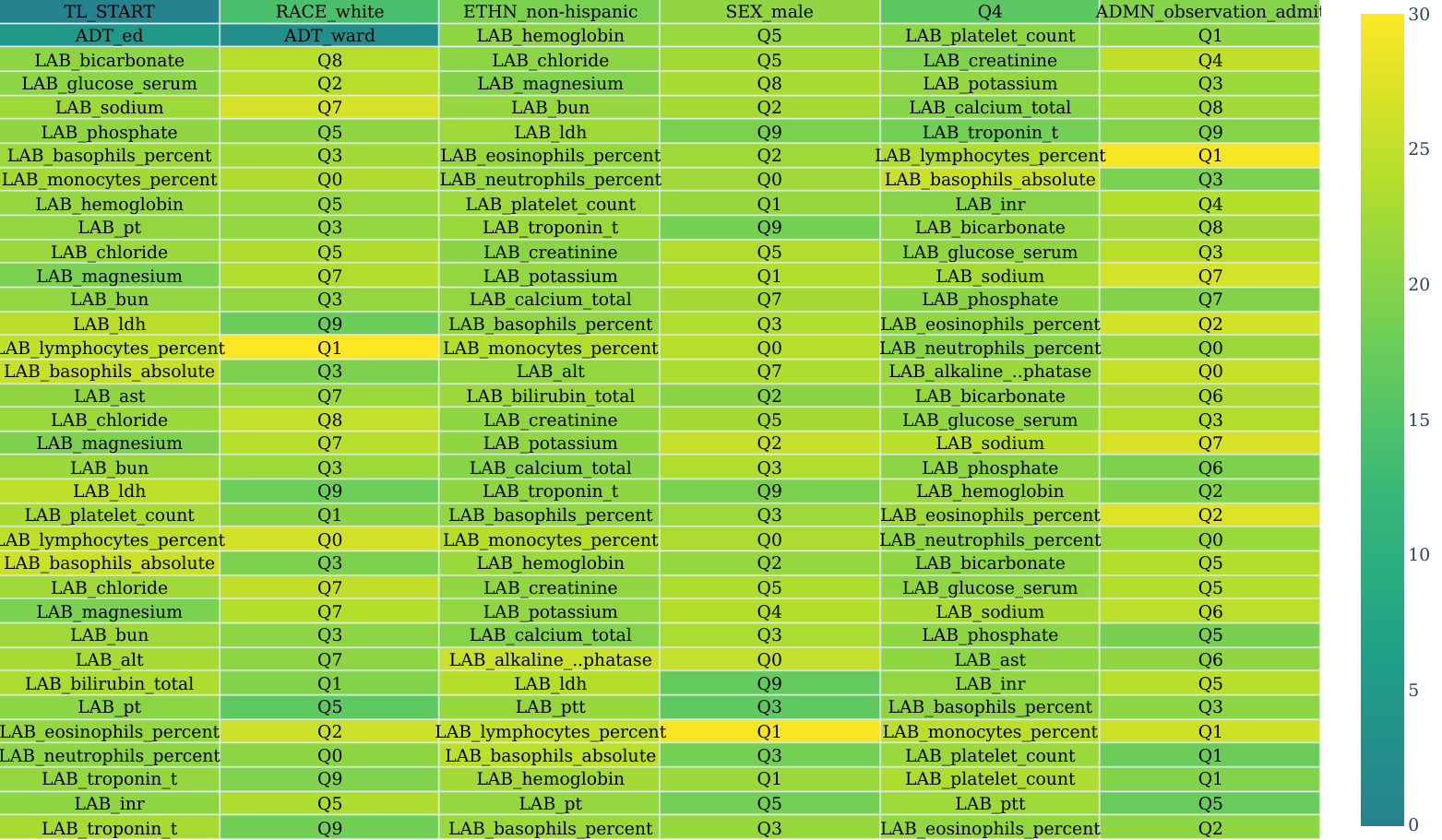}
	\caption{Timeline highlighted by tokenwise context-aware information for MIMIC hospitalization 24640534 (first 210 tokens). This white, non-Hispanic male of age $\sim 60$ was admitted to the ED for observation. He had no previous admission history within MIMIC. His stay was 37 days 22 hours in duration. After the stay, he subsequently received ICD-10 diagnoses C9200 for `Acute myeloblastic leukemia, not having achieved remission' and I214 for `Non-ST elevation (NSTEMI) myocardial infarction', among others. (Full diagnostic list in appendix.) The model successfully identified low lymphocytes as being potentially clinically relevant (\texttt{Lab\_lymphocytes\_percent}, \texttt{Q1}), but overlooked the low neutrophils (\texttt{Lab\_neutrophils\_percent}, \texttt{Q0}) and high troponin T (\texttt{Lab\_troponin\_t}, \texttt{Q9}).}
	\label{fig:mimic24640534}
\end{figure}

\begin{figure}[htb]
	\centering
	\includegraphics[width=\textwidth]{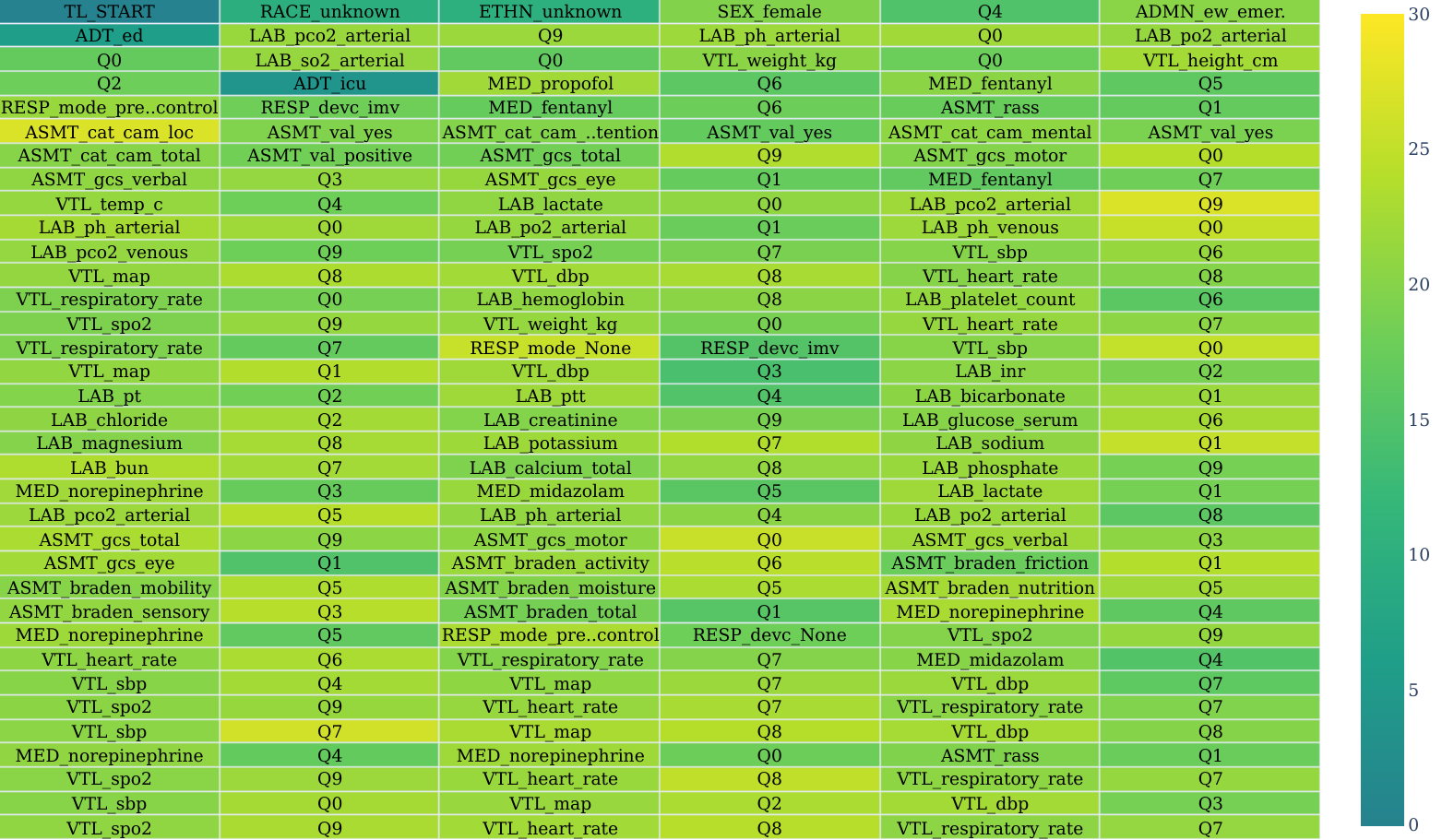}
	\caption{Timeline highlighted by tokenwise context-aware information for MIMIC hospitalization 26886976 (first 210 tokens). This female of unknown race and ethnicity was admitted to the ED at age $\sim 73$. Her 12 day 2 hour hospital stay ended in death. After the stay, she subsequently received ICD-10 diagnoses A4189 for `Other specified sepsis', R6521 for `Severe sepsis with septic shock', and N179 for `Acute kidney failure, unspecified', in addition to other diagnoses (full list in appendix). After the patient received a low Richmond Agitation-Sedation Scale score~\cite[RASS]{Ses02} with (\texttt{ASMT\_rass}, \texttt{Q1}) indicating a high likelihood of coma, the model finds the administration of the Confusion Assessment Method~\cite[CAM]{Ino90} (\texttt{ASMT\_cat\_cam\_loc}) to evaluate delirium to be surprising/ informative. The model also finds the high arterial $\text{PCO}_2$ (\texttt{LAB\_pco2\_arterial}, \texttt{Q9}) to be of interest.}
	\label{fig:mimic26886976}
\end{figure}

\begin{figure}[htb]
	\centering
	\includegraphics[width=\textwidth]{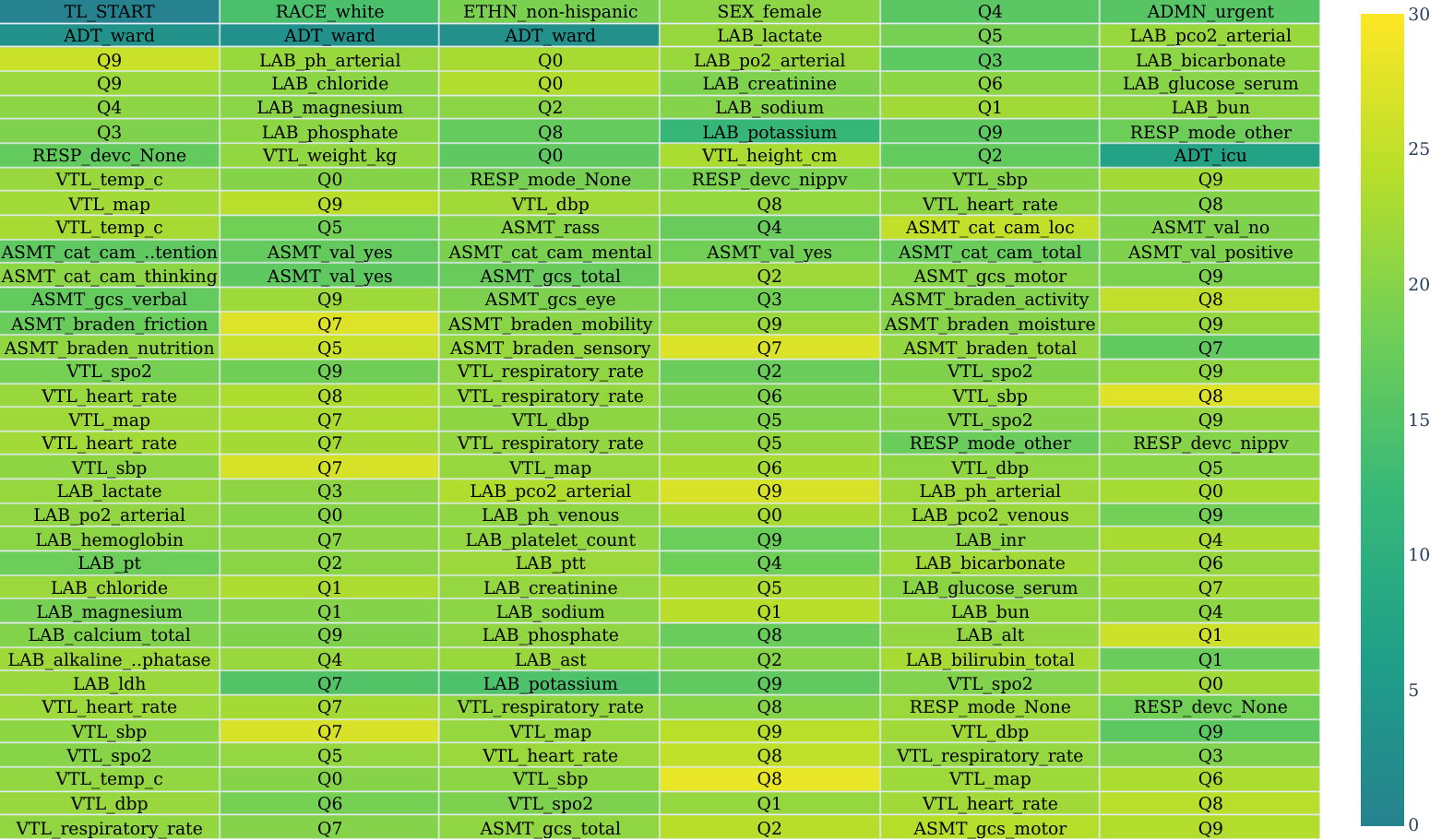}
	\caption{Timeline highlighted by tokenwise context-aware information for MIMIC hospitalization 29022625 (first 210 tokens). This $\sim 55$ year old white female had previously been seen for  a myriad of conditions (see appendix). After a a 30 day 20 hour stay, she received new diagnoses of  K2211 for `Ulcer of esophagus with bleeding',  J9601 for `Acute respiratory failure with hypoxia',  J9602 for `Acute respiratory failure with hypercapnia',  A419 for `Sepsis, unspecified organism',  J90 for `Pleural effusion, not elsewhere classified',  E872 for `Acidosis',  and J95851 for `Ventilator associated pneumonia',  among others. The model notices that the Braden scores seem implausibly high, and highlights the hypercarbia (\texttt{LAB\_pco2\_arterial}, \texttt{Q9}). It seems to miss the hypoxia (\texttt{VTL\_spo2}, \texttt{Q0}); however, $\text{SPO}_2$ readings up to 93.0 are placed in decile \texttt{Q0} so this may be due to the tokenization strategy. The model also emphasizes the patient's rising blood pressure  (\texttt{VTL\_sbp}) over time.}
	\label{fig:mimic29022625}
\end{figure}

We present the first 210 tokens from three timelines along with comments as Figures~\ref{fig:mimic24640534}-\ref{fig:mimic29022625}. We see that informative tokens sometimes correspond to lab events or vitals readings that have a direct bearing on the patient's current state. Examples include the lymphocytes percentage lab in Figure~\ref{fig:mimic24640534}, arterial $\text{PCO}_2$ in Figure~\ref{fig:mimic26886976}, and blood pressure readings in Figure~\ref{fig:mimic29022625}. Informative tokens can also correspond to clinician-initiated events, such as the CAM assessment~\cite{Ino90} in Figure~\ref{fig:mimic26886976} following a low RASS~\cite{Ses02} score. (If the RASS score were -4 as opposed to -3, typically the CAM assessment would not be made until later.) Finally, informative tokens can correspond to measurements that seem implausible and may be worth further investigation, such as the Braden scores~\cite{Ber87} in Figure~\ref{fig:mimic29022625}.

\subsection{Counts of highly informative tokens and events anticipate negative outcomes}

We consider $T_{\ge 95}$, the number of tokens exceeding the 95th percentile for informativeness, $E_{\ge 95, <99}$, the number of events in the 95th to the 99th percentile, and $E_{\ge 99}$, the number of events exceeding the 99th percentile for informativeness. For these definitions, we restrict to tokens and events that occur within the first 24 hours of admission.

In a logistic regression model for inpatient mortality in the MIMIC test set, we find that $T_{\ge 95}$ ($\hat\beta=0.0269, p<0.001$), $E_{\ge 95, <99}$ ($\hat\beta=0.3015, p<0.001$), and $E_{\ge 99}$ ($\hat\beta=0.2480, p<0.001$) all have positive coefficients and are highly significant. Similarly for long length of stay, we find that $T_{\ge 95}$ ($\hat\beta=0.0163, p<0.001$), $E_{\ge 95, <99}$ ($\hat\beta=0.2872, p<0.001$), and $E_{\ge 99}$ ($\hat\beta=0.1236, p<0.001$) are positively and significantly associated. In the UCMC test dataset (where percentiles are based on statistics from the UCMC data), $T_{\ge 95}$ ($\hat\beta=0.0148, p<0.001$), $E_{\ge 95, <99}$ ($\hat\beta=0.1684, p<0.001$), and $E_{\ge 99}$ ($\hat\beta=0.4798, p<0.001$) all associate with inpatient mortality. Similarly, $T_{\ge 95}$ ($\hat\beta=0.0165, p<0.001$), $E_{\ge 95, <99}$ ($\hat\beta=0.1292, p<0.001$), and $E_{\ge 99}$ ($\hat\beta=0.4727, p<0.001$) associate positively with long length of stay.

\subsection{Informative tokens tend to result in larger changes to a patient's latent representation}

\begin{figure}[htb]
	\centering
	\begin{minipage}[t]{.45\textwidth}
		\centering
		\includegraphics[width=\textwidth]{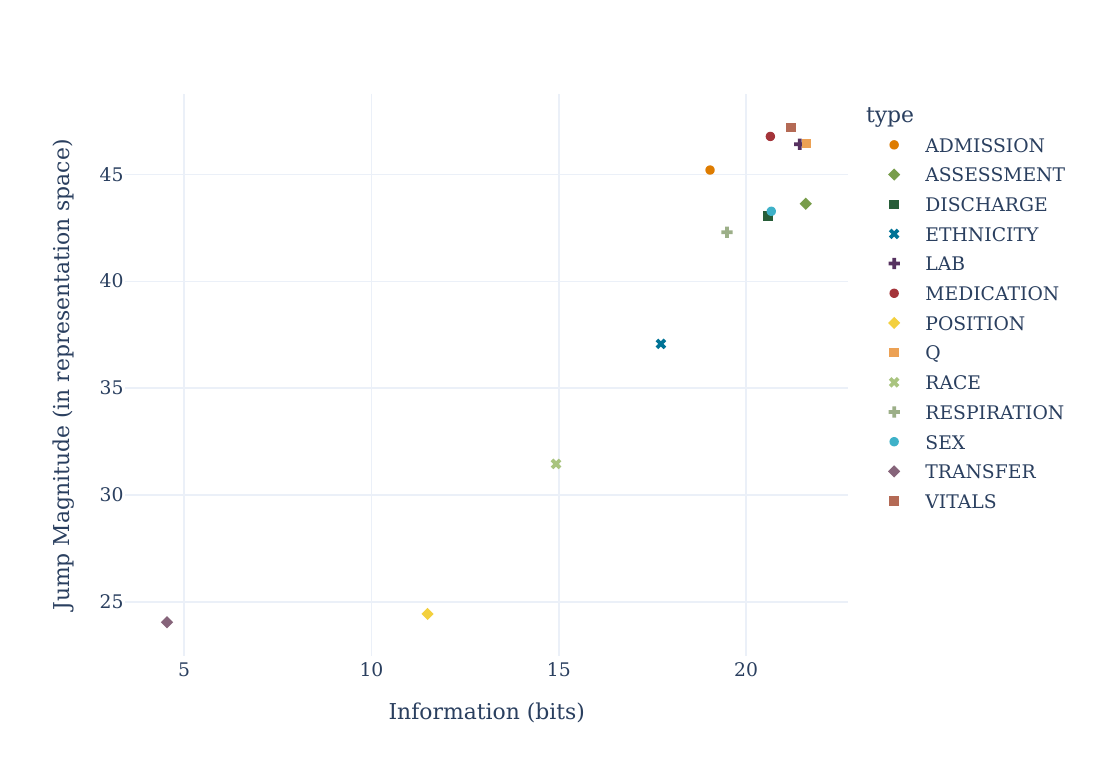}
		\caption{Average $\Delta_t$ versus average tokenwise informativeness by token type for the 24.7 million tokens $x_t$ in the MIMIC test set.}
		\label{fig:twise-jumps}
	\end{minipage}%
	\hfill
	\begin{minipage}[t]{0.45\textwidth}
		\centering
		\includegraphics[width=\textwidth]{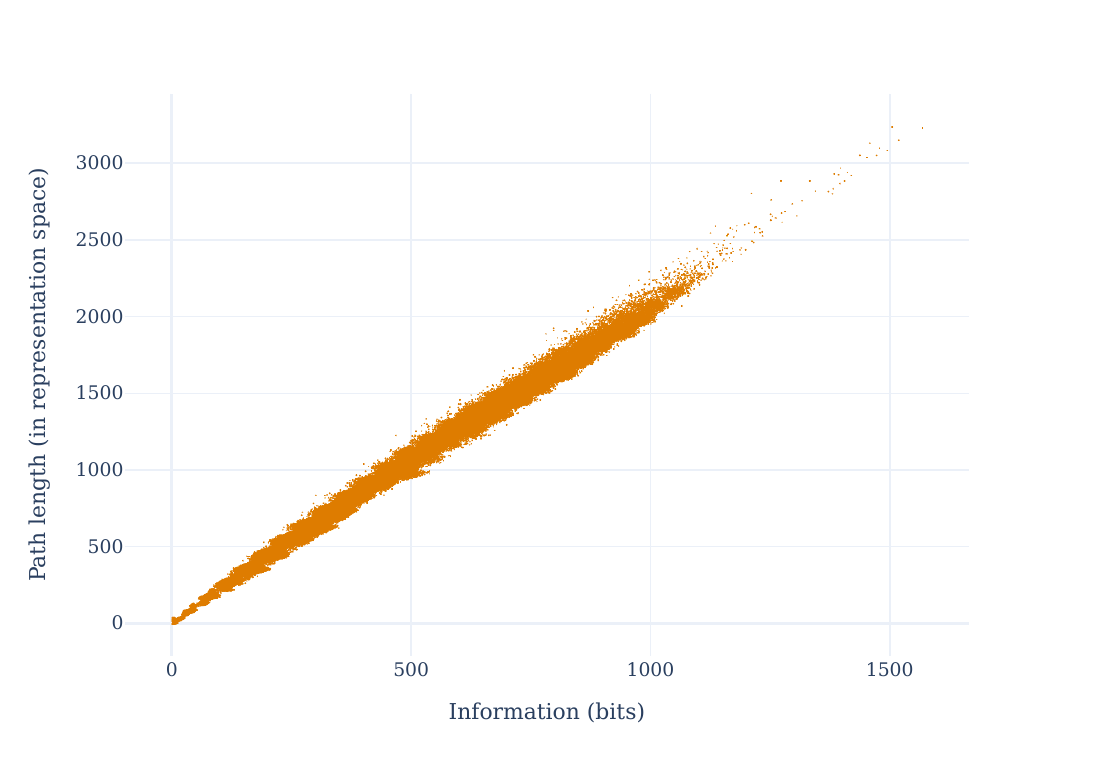}
		\caption{Path length $\Delta_{u:v}$ versus eventwise informativeness for the 2.8 million events $x_{u:v}$ in the MIMIC test set.}
		\label{fig:ewise-paths}
	\end{minipage}
\end{figure}

In our MIMIC test set, a simple linear regression for $\Delta_t$ from~\eqref{eq:delta_t} given informativeness $I_p(x_t|x_{<t})$ yields $\hat\beta=0.548, p<0.001$ with $R^2=0.212$. For a breakdown of average $\Delta_t$ vs. average informativeness by token type, see Figure~\ref{fig:twise-jumps}. Positioning and transfer tokens tend to carry less information, while assessment, lab, and quantile \texttt{Q} tokens tend to carry more.

At the level of events $x_{u:v}$, a regression for path length $\Delta_{u:v}$ from~\eqref{eq:delta_uv} given event-level informativeness $I_p(x_{u:v}|x_{<u})$ yields $\hat\beta=2.081, p<0.001$ with $R^2=0.997$. More informative events trace out longer paths in representation space. Perhaps surprisingly, there does not appear to be a strong linear relationship between event informativeness and total distance moved in representation space during the course of an event $x_{u:v}$, i.e. $\norm{R(x_{1:v})-R(x_{1:u-1})}$.

\subsection{Redacting informative events significantly reduces prognostic ability}
\label{ss:redaction_results}

Results from our redaction experiment (described in \S\ref{ss:redaction}) indicate that dropping highly informative events from a patient's timeline significantly impairs representation-based classifier performance in the MIMIC test set. For ROC-AUC, we find statistically significant performance disparities when dropping as few as 20\% of the most informative events. Conversely, events carrying little information can readily be dropped from a timeline without significantly impacting predictive performance. For representation-based logistic regression models trained on the MIMIC training set, predictive performance on the MIMIC and UCMC test sets are available in Table~\ref{tbl:auc_perf}. In addition to ROC-AUC, we report PR-AUC, the area under the precision-recall curve, and the Brier score~\cite{Bri50}, which corresponds to the mean squared error between predicted probabilities and boolean realizations in Appendix~\ref{s:full_results}. Higher ROC-AUC and PR-AUC values and lower Brier scores correspond to better performing models.

\section{Discussion}
In this work, we developed a method to quantify the informativeness of clinical events as observed in EHRs based on their tokenized representation\footnote{Shannon famously estimated the average information content of words in written English to be around 11.82 bits~\cite[Eq. 7]{Sha51}. Under the assumption that our model $p$ trained on the MIMIC training set adequately approximates the empirical distribution $\hat p$, we can average $I_p(x_t|x_{<t})$ over $x_t$ in the MIMIC test set to roughly approximate that tokens in our timelines carry around 21.23 bits of information on average.}. We found that highly informative tokens can correspond to measurements of clinical significance, to clinician-initiated events or lab orders, and in some cases to records that seem \emph{prima facie} to be data-entry errors~\cite{Niu24}. Tokens that carry more information tend to precipitate larger changes in a patient's latent representation and events that carry more information tend to have longer paths in representation space. Counts of highly informative / surprising tokens and events in the first 24 hours of a patient's stay relate to an increased risk of future negative outcomes like death or long length-of-stay. Redacting highly informative events reduces the predictive performance of representation-based classifiers, while redacting a fraction of relatively uninformative events tends to not result in significant performance drops.

\begin{table}[htb]
	\tbl{ROC-AUC for the two classification tasks on ICU patients in the MIMIC and UCMC test sets. Stars correspond to the significance level of a hypothesis test against the one-sided alternative that the model trained on the original data performs better. A single asterisk \texttt{*} corresponds to $p<0.05$, two \texttt{**} to $p<0.01$ and three \texttt{***} to $p<0.001$.}
	{
		\begin{tabular}{llllll}
			\toprule
			\multicolumn{2}{c}{version} & \multicolumn{2}{c}{Inpatient mortality} & \multicolumn{2}{c}{Long length-of-stay}                                                                                                    \\
			\cmidrule(lr){3-4} \cmidrule(lr){5-6}
			method                      & pct.                                    & MIMIC                                   & UCMC                          & MIMIC                            & UCMC                          \\ \midrule
			original                    & ---                                     & $0.869 \pm 0.009$                       & $0.839 \pm 0.013$             & $0.740 \pm 0.008$                & $0.661 \pm 0.011$             \\ \midrule
			\multirow{4}{*}{top}        & 10                                      & $0.860 \pm 0.010$                       & $0.830 \pm 0.013$             & $0.735 \pm 0.009$                & $0.671 \pm 0.011$             \\
			                            & 20                                      & $0.848 \pm 0.011$ \texttt{**}           & $0.814 \pm 0.014$ \texttt{**} & $0.726 \pm 0.009$  \texttt{**}   & $0.653 \pm 0.012$             \\
			                            & 30                                      & $0.833 \pm 0.012$ \texttt{***}          & $0.812 \pm 0.013$ \texttt{**} & $0.720 \pm 0.009$   \texttt{**}  & $0.642 \pm 0.011$  \texttt{*} \\
			                            & 40                                      & $0.823 \pm 0.011$ \texttt{***}          & $0.818 \pm 0.012$  \texttt{*} & $0.714 \pm 0.009$   \texttt{***} & $0.649 \pm 0.012$             \\ \midrule
			\multirow{4}{*}{bottom}     & 10                                      & $0.867 \pm 0.010$                       & $0.834 \pm 0.011$             & $0.736 \pm 0.012$                & $0.659 \pm 0.013$             \\
			                            & 20                                      & $0.866 \pm 0.009$                       & $0.834 \pm 0.012$             & $0.732 \pm 0.010$                & $0.667 \pm 0.012$             \\
			                            & 30                                      & $0.862 \pm 0.011$                       & $0.829 \pm 0.012$             & $0.726 \pm 0.009$  \texttt{*}    & $0.667 \pm 0.013$             \\
			                            & 40                                      & $0.859 \pm 0.012$                       & $0.829 \pm 0.011$             & $0.724 \pm 0.008$ \texttt{**}    & $0.664 \pm 0.011$             \\ \midrule
			\multirow{4}{*}{random}     & 10                                      & $0.866 \pm 0.008$                       & $0.838 \pm 0.012$             & $0.737 \pm 0.006$                & $0.664 \pm 0.012$             \\
			                            & 20                                      & $0.863 \pm 0.008$                       & $0.835 \pm 0.011$             & $0.733 \pm 0.009$                & $0.667 \pm 0.012$             \\
			                            & 30                                      & $0.865 \pm 0.010$                       & $0.835 \pm 0.014$             & $0.728 \pm 0.007$                & $0.674 \pm 0.009$             \\
			                            & 40                                      & $0.861 \pm 0.011$                       & $0.835 \pm 0.011$             & $0.727 \pm 0.008$ \texttt{*}     & $0.674 \pm 0.011$             \\
			\bottomrule
		\end{tabular}
	}
	\label{tbl:auc_perf}
\end{table}

\subsection{Broad applicability beyond clinical prediction}

The foundation model-derived informativeness measure extends well beyond traditional clinical prediction tasks and opens new avenues for AI applications to healthcare. This context-aware information quantification provides a principled framework for addressing downstream challenges that have historically relied on heuristic approaches.

Our informativeness metric could provide a data-driven solution to clinical alarm fatigue by implementing dynamic alerting systems that prioritize notifications based on contextual surprise rather than static thresholds. For example, a blood pressure of 118/86 mmHg might be routine in most contexts, but could be highly informative if it represents a rapid drop in a patient being treated for an ischemic stroke or hypertensive emergency. In these conditions, a slower reduction in blood pressure is preferred to avoid complications, making a rapid drop worthy of alerting a clinician. Traditional rule-based clinical decision support alerts would likely consider this a normal reading and therefore fail to identify a potentially concerning change. This ability to interpret context is what demonstrates the value of foundation model approaches~\cite{How24}, as they can differentiate this alarming event from a similar, but clinically appropriate, change in another patient.

The ability to detect data entry errors represents another valuable application. Traditional validation of data entry quality is dedicated to verification of abnormal values, but our approach could identify contextually implausible entries that fall within normal ranges. Overall, surprise quantification could help automate much of the manual chart review process currently required for quality assurance.

For clinical research, informativeness patterns could enable novel patient phenotyping approaches. Rather than relying on pre-specified diagnostic codes, researchers could identify patient subgroups characterized by similar patterns of surprising events, potentially revealing previously unrecognized disease subtypes or complex pathologies difficult to capture with traditional algorithms. So-called events-based models~\cite{Fon12,You14} have already proven remarkably effective at both subtyping and staging neurodegenerative disease~\cite{You18,Arc21,Vog21}, but could in the future find broader applications.

At the health system level, patterns of informativeness could inform resource allocation decisions. Units characterized by high rates of surprising events might require additional staffing or monitoring capabilities. Our finding that surprising events correlate with negative outcomes suggests informativeness patterns could serve as early warning indicators for periods of increased clinical risk.

Finally, cases highlighted by our informativeness measure could serve as valuable educational resources. Clinical scenarios with high-information events represent situations where standard protocols might be insufficient, making them ideal for training clinicians to recognize complex presentations.

Quantifying the information in EHRs could help clinicians quickly detect anomalous events and identify data entry errors. FMs trained on tokenized EHRs have already demonstrated remarkably good performance on prognostic tasks. Finding ways to better understand and interpret predictions made from these models through informativeness quantification represents a significant step toward more transparent and actionable clinical AI systems.

The broad applicability of this informativeness framework across diverse healthcare challenges suggests that context-aware information quantification could become a foundational tool for healthcare analytics, complementing traditional approaches with a more nuanced understanding of clinical surprise and significance.

\subsection*{Data and code availability}
The MIMIC-IV-3.1 dataset~\cite{Joh23} is available to credentialed users on Physionet~\cite{Gol00}. The UCMC dataset is available in the CLIF format for federated, privacy-preserving analyses. Reasonable requests may be directed to WFP. Code to reproduce the results found in this manuscript is available on Github\footnote{See: \url{https://github.com/bbj-lab/Quantifying-Surprise-EHRs}}. The appendix contains: (A) a list of all tokens used in our vocabulary, (B) the cutoffs for each category that received category-value tokenization, (C) 5 example highlighted tokenized timeline snippets from MIMIC along with subsequent diagnostic outcomes, (D) 5 examples from UCMC, and (E) a complete listing of results from our redaction experiment.

\ifarxiv
	\subsection*{Acknowledgments}

	This work was funded in part by the National Institutes of Health, specifically the National Institute of Neurological Disorders and Stroke grant number R00NS114850 to BKB and National Library of Medicine grant number R01LM014263
to WFP. This project would not have been possible without the support of the Center for Research Informatics at the University of Chicago and particularly the High-Performance Computing team led by Mike Jarsulic. The authors are grateful for the resources and support this team provided throughout the duration of the project. The Center for Research Informatics is funded by the Biological Sciences Division at the University of Chicago with additional funding provided by the Institute for Translational Medicine, CTSA grant number UL1 TR000430 from the National Institutes of Health.
\fi

\ifarxiv
\else
	\clearpage
\fi
\bibliographystyle{abbrv} 
\bibliography{detecting-informative-events}

\ifarxiv

	\newpage
	\appendix{Vocabulary}

	Our vocabulary consisted of 208 tokens, broken down by token type as follows:

	\begin{raggedright}

		\subsubsection*{Deciles (10)}
		\texttt{Q0}, \texttt{Q1}, \texttt{Q2}, \texttt{Q3}, \texttt{Q4}, \texttt{Q5}, \texttt{Q6}, \texttt{Q7}, \texttt{Q8}, \texttt{Q9}

		\subsubsection*{Special (6)}
		\texttt{TL\_START}, \texttt{TL\_END}, \texttt{PAD}, \texttt{TRUNC}, \texttt{None}, \texttt{nan}

		\subsubsection*{Race (7)}
		\texttt{RACE\_white}, \texttt{RACE\_unknown}, \texttt{RACE\_other}, \texttt{RACE\_black\_or\_african\_american}, \texttt{RACE\_asian}, \texttt{RACE\_american\_indian\_or\_alaska\_native}, \texttt{RACE\_native\_hawaiian\_or\_other\_pacific\_islander}

		\subsubsection*{Ethnicity (3)}
		\texttt{ETHN\_non-hispanic}, \texttt{ETHN\_unknown}, \texttt{ETHN\_hispanic}

		\subsubsection*{Sex (2) }
		\texttt{SEX\_female}, \texttt{SEX\_male}

		\subsubsection*{Admission (9)}
		\texttt{ADMN\_ew\_emer.}, \texttt{ADMN\_eu\_observation}, \texttt{ADMN\_urgent}, \texttt{ADMN\_surgical\_same\_day\_admission}, \texttt{ADMN\_direct\_emer.}, \texttt{ADMN\_direct\_observation}, \texttt{ADMN\_ambulatory\_observation}, \texttt{ADMN\_observation\_admit}, \texttt{ADMN\_elective}

		\subsubsection*{Discharge (12)}
		\texttt{DSCG\_hospice}, \texttt{DSCG\_missing}, \texttt{DSCG\_acute\_inpatient\_rehab\_facility}, \texttt{DSCG\_home}, \texttt{DSCG\_expired}, \texttt{DSCG\_other}, \texttt{DSCG\_skilled\_nursing\_facility\_(snf)}, \texttt{DSCG\_against\_medical\_advice\_(ama)}, \texttt{DSCG\_long\_term\_care\_hospital\_(ltach)}, \texttt{DSCG\_acute\_care\_hospital}, \texttt{DSCG\_psychiatric\_hospital}, \texttt{DSCG\_assisted\_living}

		\subsubsection*{Transfer (8)}
		\texttt{ADT\_ed}, \texttt{ADT\_ward}, \texttt{ADT\_icu}, \texttt{ADT\_l\&d}, \texttt{ADT\_psych}, \texttt{ADT\_stepdown}, \texttt{ADT\_other}, \texttt{ADT\_procedural}

		\subsubsection*{Labs (45)}
		\texttt{LAB\_hemoglobin}, \texttt{LAB\_platelet\_count}, \texttt{LAB\_bicarbonate}, \texttt{LAB\_chloride}, \texttt{LAB\_creatinine}, \texttt{LAB\_glucose\_serum}, \texttt{LAB\_magnesium}, \texttt{LAB\_potassium}, \texttt{LAB\_sodium}, \texttt{LAB\_bun}, \texttt{LAB\_inr}, \texttt{LAB\_pt}, \texttt{LAB\_ptt}, \texttt{LAB\_basophils\_percent}, \texttt{LAB\_eosinophils\_percent}, \texttt{LAB\_lymphocytes\_percent}, \texttt{LAB\_monocytes\_percent}, \texttt{LAB\_neutrophils\_percent}, \texttt{LAB\_basophils\_absolute}, \texttt{LAB\_albumin}, \texttt{LAB\_ferritin}, \texttt{LAB\_troponin\_t}, \texttt{LAB\_calcium\_total}, \texttt{LAB\_phosphate}, \texttt{LAB\_alt}, \texttt{LAB\_alkaline\_phosphatase}, \texttt{LAB\_ast}, \texttt{LAB\_bilirubin\_total}, \texttt{LAB\_ldh}, \texttt{LAB\_lactate}, \texttt{LAB\_pco2\_arterial}, \texttt{LAB\_ph\_arterial}, \texttt{LAB\_po2\_arterial}, \texttt{LAB\_bilirubin\_conjugated}, \texttt{LAB\_bilirubin\_unconjugated}, \texttt{LAB\_total\_protein}, \texttt{LAB\_calcium\_ionized}, \texttt{LAB\_so2\_arterial}, \texttt{LAB\_crp}, \texttt{LAB\_esr}, \texttt{LAB\_wbc}, \texttt{LAB\_ph\_venous}, \texttt{LAB\_pco2\_venous}, \texttt{LAB\_so2\_mixed\_venous}, \texttt{LAB\_so2\_central\_venous}

		\subsubsection*{Vitals (9)}
		\texttt{VTL\_spo2}, \texttt{VTL\_sbp}, \texttt{VTL\_map}, \texttt{VTL\_weight\_kg}, \texttt{VTL\_dbp}, \texttt{VTL\_heart\_rate}, \texttt{VTL\_respiratory\_rate}, \texttt{VTL\_height\_cm}, \texttt{VTL\_temp\_c}

		\subsubsection*{Medicines (46)}
		\texttt{MED\_dextrose}, \texttt{MED\_dobutamine}, \texttt{MED\_norepinephrine}, \texttt{MED\_vasopressin}, \texttt{MED\_phenylephrine}, \texttt{MED\_magnesium}, \texttt{MED\_propofol}, \texttt{MED\_insulin}, \texttt{MED\_octreotide}, \texttt{MED\_epinephrine}, \texttt{MED\_pantoprazole}, \texttt{MED\_morphine}, \texttt{MED\_nicardipine}, \texttt{MED\_fentanyl}, \texttt{MED\_sodium bicarbonate}, \texttt{MED\_diltiazem}, \texttt{MED\_dexmedetomidine}, \texttt{MED\_amiodarone}, \texttt{MED\_heparin}, \texttt{MED\_midazolam}, \texttt{MED\_cisatracurium}, \texttt{MED\_hydromorphone}, \texttt{MED\_tpn}, \texttt{MED\_milrinone}, \texttt{MED\_eptifibatide}, \texttt{MED\_dopamine}, \texttt{MED\_argatroban}, \texttt{MED\_lidocaine}, \texttt{MED\_furosemide}, \texttt{MED\_rocuronium}, \texttt{MED\_vecuronium}, \texttt{MED\_pentobarbital}, \texttt{MED\_esmolol}, \texttt{MED\_labetalol}, \texttt{MED\_nitroprusside}, \texttt{MED\_angiotensin}, \texttt{MED\_ketamine}, \texttt{MED\_clevidipine}, \texttt{MED\_lorazepam}, \texttt{MED\_bumetanide}, \texttt{MED\_naloxone}, \texttt{MED\_procainamide}, \texttt{MED\_aminocaproic}, \texttt{MED\_aminophylline}, \texttt{MED\_treprostinil}, \texttt{MED\_epoprostenol}

		\subsubsection*{Assessments (32)}
		\texttt{ASMT\_gcs\_total}, \texttt{ASMT\_gcs\_motor}, \texttt{ASMT\_gcs\_verbal}, \texttt{ASMT\_gcs\_eye}, \texttt{ASMT\_rass}, \texttt{ASMT\_braden\_activity}, \texttt{ASMT\_braden\_friction}, \texttt{ASMT\_braden\_mobility}, \texttt{ASMT\_braden\_moisture}, \texttt{ASMT\_braden\_nutrition}, \texttt{ASMT\_braden\_sensory}, \texttt{ASMT\_braden\_total}, \texttt{ASMT\_cat\_cam\_loc}, \texttt{ASMT\_cat\_cam\_inattention}, \texttt{ASMT\_cat\_cam\_mental}, \texttt{ASMT\_cat\_cam\_total}, \texttt{ASMT\_cat\_cam\_thinking}, \texttt{ASMT\_val\_yes}, \texttt{ASMT\_val\_positive}, \texttt{ASMT\_val\_no}, \texttt{ASMT\_val\_negative}, \texttt{ASMT\_val\_unable\_to\_assess}, \texttt{ASMT\_val\_no\_(stop\_-\_not\_delirious)}, \texttt{ASMT\_val\_language\_barrier}, \texttt{ASMT\_val\_preexisting\_advanced\_dementia}, \texttt{ASMT\_val\_yes\_(continue)}, \texttt{ASMT\_val\_unable\_to\_assess\_(stop)}, \texttt{ASMT\_val\_yes\_(3\_or\_more\_errors,\_then\_continue)}, \texttt{ASMT\_val\_no\_(less\_than\_3\_errors\_-\_stop\_-\_not\_delirious)}, \texttt{ASMT\_cat\_sbt\_delivery\_pass\_fail}, \texttt{ASMT\_val\_pass}, \texttt{ASMT\_val\_fail}

		\subsubsection*{Respiratory (18)}
		\texttt{RESP\_mode\_None}, \texttt{RESP\_mode\_assist\_control-volume\_control}, \texttt{RESP\_mode\_pressure\_support/cpap}, \texttt{RESP\_mode\_pressure-regulated\_volume\_control}, \texttt{RESP\_mode\_other}, \texttt{RESP\_mode\_volume\_support}, \texttt{RESP\_mode\_simv}, \texttt{RESP\_mode\_blow\_by}, \texttt{RESP\_mode\_pressure\_control}, \texttt{RESP\_devc\_nasal\_cannula}, \texttt{RESP\_devc\_imv}, \texttt{RESP\_devc\_None}, \texttt{RESP\_devc\_face\_mask}, \texttt{RESP\_devc\_high\_flow\_nc}, \texttt{RESP\_devc\_nippv}, \texttt{RESP\_devc\_trach\_collar}, \texttt{RESP\_devc\_other}, \texttt{RESP\_devc\_cpap}

		\subsubsection*{Positioning (1)}
		\texttt{POSN\_prone}

	\end{raggedright}

	\newpage

	\appendix{Decile cutoffs}
	This section contains tables corresponding to the 9 cutoffs for the 10 deciles. Values corresponding to each category were binned using the cutoffs $C1 \leq C2 \leq C3 \leq \dotsb \leq C9$. Values in $(-\infty,C_1)$ were assigned to \texttt{Q0}, values in $[C_1,C_2)$ to \texttt{Q1}, values in $[C_2, C_3)$ to \texttt{Q2}, and so on, up to values in $[C_9, \infty)$, which were assigned to \texttt{Q9}.

	\begin{table}[htb]
		\centering
		\small
		\tbl{Decile cutoffs for age at admission.}
		{
			\begin{tabular}{lrrrrrrrrr}
				\toprule
				category           & $C_1$ & $C_2$ & $C_3$ & $C_4$ & $C_5$ & $C_6$ & $C_7$ & $C_8$ & $C_9$ \\ \midrule
				age\_at\_admission & 30.0  & 40.0  & 49.0  & 55.0  & 61.0  & 66.0  & 71.0  & 77.0  & 84.0  \\
				\bottomrule
			\end{tabular}
		}
	\end{table}

	\begin{table}[htb]
		\centering
		\small
		\tbl{Decile cutoffs for vitals.}
		{
			\begin{tabular}{lrrrrrrrrr} \toprule
				category               & $C_1$ & $C_2$ & $C_3$ & $C_4$ & $C_5$ & $C_6$ & $C_7$ & $C_8$ & $C_9$ \\ \midrule
				VTL\_spo2              & 93.0  & 95.0  & 96.0  & 96.0  & 97.0  & 98.0  & 99.0  & 100.0 & 100.0 \\
				VTL\_sbp               & 93.0  & 100.0 & 106.0 & 111.0 & 117.0 & 123.0 & 129.0 & 138.0 & 150.0 \\
				VTL\_map               & 61.0  & 66.0  & 70.0  & 73.0  & 77.0  & 81.0  & 85.0  & 91.0  & 100.0 \\
				VTL\_weight\_kg        & 58.2  & 65.9  & 71.6  & 77.0  & 82.7  & 88.7  & 95.3  & 103.7 & 116.7 \\
				VTL\_dbp               & 46.0  & 50.0  & 54.0  & 58.0  & 61.0  & 65.0  & 69.0  & 74.0  & 83.0  \\
				VTL\_heart\_rate       & 64.0  & 70.0  & 76.0  & 80.0  & 85.0  & 90.0  & 95.0  & 101.0 & 111.0 \\
				VTL\_respiratory\_rate & 13.0  & 15.0  & 17.0  & 18.0  & 19.0  & 21.0  & 23.0  & 25.0  & 28.0  \\
				VTL\_height\_cm        & 155.0 & 160.0 & 163.0 & 168.0 & 170.0 & 173.0 & 175.0 & 178.0 & 183.0 \\
				VTL\_temp\_c           & 36.3  & 36.5  & 36.7  & 36.8  & 36.9  & 37.1  & 37.2  & 37.4  & 37.8  \\ \bottomrule
			\end{tabular}
		}
	\end{table}

	\begin{table}[htb]
		\centering
		\small
		\tbl{Decile cutoffs for assessments.}
		{
			\begin{tabular}{lrrrrrrrrr} \toprule
				category                & $C_1$ & $C_2$ & $C_3$ & $C_4$ & $C_5$ & $C_6$ & $C_7$ & $C_8$ & $C_9$ \\ \midrule
				ASMT\_gcs\_total        & 13.0  & 14.0  & 15.0  & 15.0  & 15.0  & 15.0  & 15.0  & 15.0  & 15.0  \\
				ASMT\_gcs\_motor        & 3.0   & 5.0   & 6.0   & 6.0   & 6.0   & 6.0   & 6.0   & 6.0   & 6.0   \\
				ASMT\_gcs\_verbal       & 0.0   & 0.0   & 0.0   & 1.0   & 4.0   & 5.0   & 5.0   & 5.0   & 5.0   \\
				ASMT\_gcs\_eye          & 1.0   & 3.0   & 3.0   & 4.0   & 4.0   & 4.0   & 4.0   & 4.0   & 4.0   \\
				ASMT\_rass              & -4.0  & -2.0  & -1.0  & -1.0  & 0.0   & 0.0   & 0.0   & 0.0   & 1.0   \\
				ASMT\_braden\_activity  & 1.0   & 1.0   & 1.0   & 1.0   & 1.0   & 1.0   & 2.0   & 2.0   & 3.0   \\
				ASMT\_braden\_friction  & 1.0   & 2.0   & 2.0   & 2.0   & 2.0   & 2.0   & 2.0   & 3.0   & 3.0   \\
				ASMT\_braden\_mobility  & 1.0   & 2.0   & 2.0   & 2.0   & 2.0   & 3.0   & 3.0   & 3.0   & 3.0   \\
				ASMT\_braden\_moisture  & 3.0   & 3.0   & 3.0   & 3.0   & 3.0   & 4.0   & 4.0   & 4.0   & 4.0   \\
				ASMT\_braden\_nutrition & 2.0   & 2.0   & 2.0   & 2.0   & 2.0   & 3.0   & 3.0   & 3.0   & 3.0   \\
				ASMT\_braden\_sensory   & 2.0   & 2.0   & 2.0   & 3.0   & 3.0   & 3.0   & 3.0   & 4.0   & 4.0   \\
				ASMT\_braden\_total     & 11.0  & 12.0  & 13.0  & 14.0  & 14.0  & 15.0  & 16.0  & 17.0  & 19.0  \\ \bottomrule
			\end{tabular}
		}
	\end{table}

	\begin{table}[htb]
		\centering
		\small
		\tbl{Decile cutoffs for labs.}
		{
			\begin{tabular}{lrrrrrrrrr}
				\toprule
				category                     & $C_1$ & $C_2$ & $C_3$ & $C_4$ & $C_5$ & $C_6$ & $C_7$ & $C_8$  & $C_9$  \\ \midrule
				LAB\_hemoglobin              & 7.6   & 8.2   & 8.7   & 9.2   & 9.8   & 10.5  & 11.1  & 11.9   & 13.0   \\
				LAB\_platelet\_count         & 70.0  & 119.0 & 152.0 & 180.0 & 206.0 & 234.0 & 266.0 & 310.0  & 387.0  \\
				LAB\_bicarbonate             & 20.0  & 22.0  & 23.0  & 24.0  & 25.0  & 26.0  & 27.0  & 28.0   & 30.0   \\
				LAB\_chloride                & 95.0  & 98.0  & 100.0 & 101.0 & 102.0 & 104.0 & 105.0 & 107.0  & 109.0  \\
				LAB\_creatinine              & 0.5   & 0.7   & 0.7   & 0.8   & 0.9   & 1.1   & 1.3   & 1.7    & 2.8    \\
				LAB\_glucose\_serum          & 85.0  & 93.0  & 100.0 & 107.0 & 115.0 & 125.0 & 138.0 & 157.0  & 194.0  \\
				LAB\_magnesium               & 1.7   & 1.8   & 1.9   & 1.9   & 2.0   & 2.1   & 2.1   & 2.2    & 2.4    \\
				LAB\_potassium               & 3.5   & 3.7   & 3.8   & 4.0   & 4.1   & 4.2   & 4.4   & 4.6    & 4.9    \\
				LAB\_sodium                  & 132.0 & 135.0 & 136.0 & 137.0 & 138.0 & 139.0 & 140.0 & 142.0  & 144.0  \\
				LAB\_bun                     & 8.0   & 11.0  & 13.0  & 16.0  & 19.0  & 23.0  & 28.0  & 37.0   & 54.0   \\
				LAB\_inr                     & 1.0   & 1.1   & 1.2   & 1.2   & 1.3   & 1.4   & 1.5   & 1.9    & 2.4    \\
				LAB\_pt                      & 11.4  & 12.1  & 12.8  & 13.4  & 14.2  & 15.2  & 16.9  & 20.1   & 26.3   \\
				LAB\_ptt                     & 26.0  & 28.1  & 29.9  & 31.9  & 34.6  & 38.9  & 47.6  & 61.1   & 78.7   \\
				LAB\_basophils\_percent      & 0.0   & 0.0   & 0.0   & 0.2   & 0.2   & 0.3   & 0.4   & 0.6    & 1.0    \\
				LAB\_eosinophils\_percent    & 0.0   & 0.0   & 0.1   & 0.4   & 1.0   & 1.3   & 2.0   & 3.0    & 4.7    \\
				LAB\_lymphocytes\_percent    & 4.0   & 6.9   & 9.3   & 12.0  & 15.2  & 19.0  & 23.9  & 30.9   & 45.0   \\
				LAB\_monocytes\_percent      & 1.8   & 3.0   & 4.3   & 5.3   & 6.3   & 7.4   & 8.8   & 10.4   & 13.9   \\
				LAB\_neutrophils\_percent    & 30.0  & 51.7  & 60.5  & 66.7  & 71.7  & 76.0  & 80.05 & 84.3   & 89.0   \\
				LAB\_basophils\_absolute     & 0.0   & 0.0   & 0.0   & 0.01  & 0.02  & 0.03  & 0.04  & 0.05   & 0.07   \\
				LAB\_albumin                 & 2.3   & 2.6   & 2.9   & 3.0   & 3.2   & 3.4   & 3.6   & 3.8    & 4.1    \\
				LAB\_ferritin                & 43.0  & 90.0  & 149.0 & 224.0 & 325.0 & 468.0 & 692.0 & 1085.0 & 2000.0 \\
				LAB\_troponin\_t             & 10.0  & 10.0  & 10.0  & 10.0  & 20.0  & 50.0  & 100.0 & 220.0  & 670.0  \\
				LAB\_calcium\_total          & 7.8   & 8.1   & 8.3   & 8.5   & 8.7   & 8.8   & 9.0   & 9.2    & 9.5    \\
				LAB\_phosphate               & 2.3   & 2.7   & 3.0   & 3.2   & 3.4   & 3.7   & 3.9   & 4.3    & 4.9    \\
				LAB\_alt                     & 10.0  & 14.0  & 18.0  & 22.0  & 28.0  & 37.0  & 51.0  & 77.0   & 151.0  \\
				LAB\_alkaline\_phosphatase   & 54.0  & 65.0  & 76.0  & 87.0  & 101.0 & 119.0 & 145.0 & 190.0  & 296.0  \\
				LAB\_ast                     & 14.0  & 18.0  & 22.0  & 27.0  & 34.0  & 43.0  & 57.0  & 82.0   & 148.0  \\
				LAB\_bilirubin\_total        & 0.2   & 0.3   & 0.4   & 0.5   & 0.6   & 0.8   & 1.2   & 2.1    & 5.3    \\
				LAB\_ldh                     & 151.0 & 175.0 & 197.0 & 221.0 & 248.0 & 283.0 & 333.0 & 419.0  & 628.0  \\
				LAB\_lactate                 & 0.9   & 1.1   & 1.3   & 1.5   & 1.7   & 2.0   & 2.4   & 3.0    & 4.4    \\
				LAB\_pco2\_arterial          & 31.0  & 35.0  & 37.0  & 39.0  & 41.0  & 43.0  & 46.0  & 50.0   & 58.0   \\
				LAB\_ph\_arterial            & 7.26  & 7.31  & 7.34  & 7.36  & 7.38  & 7.4   & 7.42  & 7.44   & 7.47   \\
				LAB\_po2\_arterial           & 46.0  & 67.0  & 79.0  & 90.0  & 102.0 & 116.0 & 137.0 & 170.0  & 259.0  \\
				LAB\_bilirubin\_conjugated   & 0.2   & 0.3   & 0.4   & 0.7   & 1.1   & 1.8   & 2.8   & 4.6    & 8.2    \\
				LAB\_bilirubin\_unconjugated & 0.3   & 0.4   & 0.6   & 0.8   & 1.0   & 1.3   & 1.7   & 2.4    & 4.1    \\
				LAB\_total\_protein          & 4.8   & 5.2   & 5.4   & 5.7   & 5.9   & 6.1   & 6.3   & 6.6    & 7.1    \\
				LAB\_calcium\_ionized        & 4.04  & 4.2   & 4.32  & 4.4   & 4.48  & 4.56  & 4.68  & 4.76   & 4.96   \\
				LAB\_so2\_arterial           & 57.0  & 66.0  & 77.0  & 91.0  & 94.0  & 96.0  & 97.0  & 97.0   & 98.0   \\
				LAB\_crp                     & 2.7   & 7.0   & 13.8  & 30.2  & 47.7  & 69.8  & 98.9  & 141.9  & 213.92 \\
				LAB\_esr                     & 6.0   & 14.0  & 23.0  & 33.0  & 45.0  & 59.0  & 74.0  & 92.0   & 116.0  \\
				LAB\_wbc                     & 2.6   & 3.6   & 4.4   & 5.2   & 6.0   & 7.0   & 8.3   & 10.1   & 13.6   \\
				LAB\_ph\_venous              & 7.24  & 7.29  & 7.32  & 7.35  & 7.37  & 7.39  & 7.41  & 7.43   & 7.46   \\
				LAB\_pco2\_venous            & 32.0  & 36.0  & 39.0  & 42.0  & 45.0  & 48.0  & 52.0  & 58.0   & 68.0   \\
				LAB\_so2\_mixed\_venous      & 52.0  & 57.0  & 60.0  & 63.0  & 65.0  & 68.0  & 70.0  & 73.0   & 78.0   \\
				LAB\_so2\_central\_venous    & 50.0  & 57.0  & 62.0  & 67.0  & 71.0  & 74.0  & 78.0  & 81.0   & 86.0   \\
				\bottomrule
			\end{tabular}
		}
	\end{table}

	\begin{table}[htb]
		\centering
		\small
		\tbl{Decile cutoffs for medicines.}
		{
			\begin{tabular}{lrrrrrrrrr} \toprule
				category                 & $C_1$ & $C_2$ & $C_3$ & $C_4$ & $C_5$   & $C_6$   & $C_7$   & $C_8$  & $C_9$   \\ \midrule
				MED\_dextrose            & 0.0   & 2.0   & 5.18  & 8.21  & 10.65   & 14.94   & 20.78   & 31.09  & 50.03   \\
				MED\_dobutamine          & 0.0   & 2.0   & 2.5   & 2.5   & 4.27    & 5.0     & 5.01    & 7.33   & 8.05    \\
				MED\_norepinephrine      & 0.01  & 0.03  & 0.05  & 0.06  & 0.09    & 0.12    & 0.15    & 0.2    & 0.3     \\
				MED\_vasopressin         & 0.0   & 0.0   & 1.2   & 1.21  & 1.81    & 2.4     & 2.4     & 2.4    & 3.59    \\
				MED\_phenylephrine       & 0.0   & 0.27  & 0.45  & 0.5   & 0.76    & 1.0     & 1.26    & 1.86   & 2.93    \\
				MED\_magnesium           & 0.0   & 0.0   & 50.0  & 50.0  & 50.0    & 50.0    & 50.0    & 50.0   & 50.0    \\
				MED\_propofol            & 0.0   & 10.04 & 20.0  & 24.97 & 30.05   & 39.11   & 40.48   & 50.11  & 60.2    \\
				MED\_insulin             & 0.0   & 1.0   & 2.0   & 2.99  & 3.0     & 4.0     & 5.0     & 6.52   & 9.41    \\
				MED\_octreotide          & 0.0   & 0.0   & 49.98 & 50.0  & 50.0    & 50.0    & 50.02   & 50.17  & 50.35   \\
				MED\_epinephrine         & 0.0   & 0.01  & 0.02  & 0.03  & 0.04    & 0.05    & 0.08    & 0.13   & 0.3     \\
				MED\_pantoprazole        & 0.0   & 0.0   & 8.0   & 8.0   & 8.0     & 8.0     & 8.0     & 8.03   & 8.07    \\
				MED\_morphine            & 0.0   & 0.0   & 2.0   & 2.0   & 4.0     & 5.0     & 6.0     & 10.0   & 14.98   \\
				MED\_nicardipine         & 0.0   & 0.5   & 0.5   & 1.0   & 1.0     & 1.5     & 1.97    & 2.02   & 3.0     \\
				MED\_fentanyl            & 0.0   & 25.0  & 50.0  & 50.05 & 75.04   & 100.0   & 125.02  & 151.8  & 209.3   \\
				MED\_sodium\_bicarbonate & 0.0   & 0.0   & 0.0   & 0.0   & 0.0     & 0.0     & 0.0     & 0.0    & 0.0     \\
				MED\_diltiazem           & 0.0   & 4.99  & 5.0   & 5.01  & 9.99    & 10.0    & 10.05   & 14.99  & 15.02   \\
				MED\_dexmedetomidine     & 0.0   & 0.4   & 0.4   & 0.6   & 0.7     & 0.81    & 1.0     & 1.2    & 1.4     \\
				MED\_amiodarone          & 0.0   & 0.0   & 0.5   & 0.5   & 0.5     & 0.5     & 1.0     & 1.0    & 1.0     \\
				MED\_heparin             & 0.0   & 0.0   & 650.2 & 867.9 & 1030.65 & 1200.23 & 1400.31 & 1612.9 & 1985.37 \\
				MED\_midazolam           & 0.0   & 0.5   & 1.0   & 2.0   & 2.01    & 3.04    & 4.04    & 6.01   & 10.01   \\
				MED\_cisatracurium       & 0.0   & 0.06  & 0.1   & 0.13  & 0.15    & 0.18    & 0.2     & 0.25   & 0.3     \\
				MED\_hydromorphone       & 0.25  & 1.0   & 1.11  & 2.0   & 2.5     & 3.02    & 4.0     & 4.5    & 7.03    \\
				MED\_tpn                 & 0.0   & 0.0   & 0.0   & 42.1  & 58.3    & 63.95   & 72.9    & 80.83  & 91.03   \\
				MED\_milrinone           & 0.0   & 0.13  & 0.25  & 0.25  & 0.25    & 0.38    & 0.38    & 0.5    & 0.5     \\
				MED\_eptifibatide        & 0.0   & 0.0   & 0.0   & 2.0   & 2.0     & 2.0     & 2.0     & 2.0    & 2.02    \\
				MED\_dopamine            & 0.0   & 2.0   & 3.0   & 4.0   & 5.0     & 5.98    & 7.58    & 10.01  & 14.02   \\
				MED\_argatroban          & 0.0   & 0.1   & 0.5   & 0.69  & 1.0     & 1.25    & 1.72    & 2.36   & 3.44    \\
				MED\_lidocaine           & 0.0   & 0.0   & 0.5   & 1.0   & 1.0     & 1.0     & 1.0     & 2.0    & 2.0     \\
				MED\_furosemide          & 0.0   & 1.5   & 4.0   & 5.0   & 8.0     & 10.0    & 14.05   & 19.96  & 20.2    \\
				MED\_rocuronium          & 0.0   & 6.01  & 8.0   & 8.01  & 8.02    & 8.09    & 9.03    & 10.01  & 11.06   \\
				MED\_vecuronium          & 0.0   & 0.0   & 0.03  & 0.05  & 0.05    & 0.05    & 0.05    & 0.08   & 0.1     \\
				MED\_pentobarbital       & 0.0   & 0.5   & 1.0   & 1.82  & 2.12    & 3.01    & 3.95    & 4.99   & 5.24    \\
				MED\_esmolol             & 0.0   & 48.31 & 50.1  & 81.07 & 100.35  & 143.33  & 155.25  & 200.93 & 257.79  \\
				MED\_labetalol           & 0.0   & 0.5   & 0.5   & 1.0   & 1.0     & 1.5     & 2.0     & 2.5    & 3.75    \\
				MED\_nitroprusside       & 0.0   & 0.3   & 0.5   & 0.6   & 0.8     & 1.0     & 1.4     & 1.81   & 2.42    \\
				MED\_angiotensin         & 0.0   & 0.02  & 0.04  & 5.01  & 20.0    & 20.02   & 34.99   & 40.03  & 52.09   \\
				MED\_ketamine            & 0.1   & 0.2   & 0.3   & 0.4   & 0.5     & 0.63    & 0.9     & 1.14   & 1.39    \\
				MED\_clevidipine         & 0.51  & 2.0   & 2.17  & 4.0   & 4.99    & 6.02    & 8.01    & 10.04  & 14.0    \\
				MED\_lorazepam           & 0.0   & 0.5   & 1.0   & 2.0   & 2.01    & 3.0     & 4.0     & 5.0    & 6.7     \\
				MED\_bumetanide          & 0.5   & 1.0   & 2.0   & 2.0   & 2.02    & 2.21    & 3.06    & 4.0    & 4.08    \\
				MED\_naloxone            & 0.0   & 0.0   & 0.0   & 0.1   & 0.2     & 0.2     & 0.2     & 0.2    & 0.3     \\
				MED\_procainamide        & 0.0   & 0.5   & 1.5   & 2.0   & 2.02    & 4.02    & 4.08    & 5.0    & 5.0     \\
				MED\_aminocaproic        & 0.0   & 0.0   & 0.5   & 1.0   & 1.0     & 1.0     & 1.01    & 4.0    & 1000.0  \\
				MED\_aminophylline       & 0.21  & 0.3   & 0.3   & 0.3   & 0.3     & 0.3     & 0.3     & 0.36   & 0.4     \\
				MED\_treprostinil        & 0.0   & 0.03  & 0.78  & 6.0   & 9.02    & 11.12   & 14.16   & 17.0   & 21.15   \\
				MED\_epoprostenol        & 0.05  & 6.05  & 29.07 & 42.0  & 42.0    & 42.01   & 42.02   & 42.14  & 42.26   \\ \bottomrule
			\end{tabular}
		}
	\end{table}

    \clearpage
	\appendix{Example timelines from MIMIC}

	\begin{figure}[htb]
		\centering
		\includegraphics[width=0.8\textwidth]{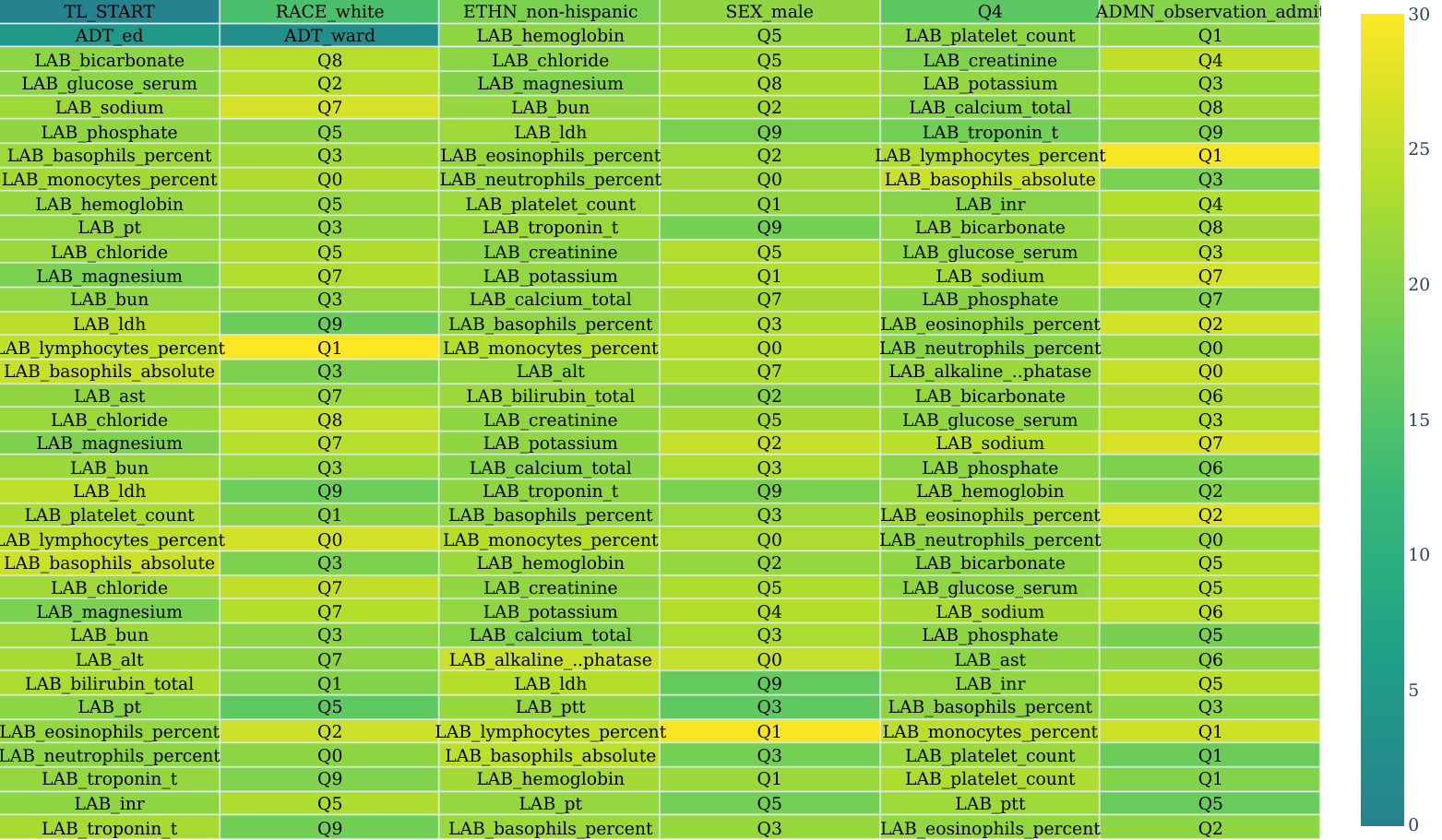}
		\caption{Tokenwise context-aware information for the first 210 tokens of MIMIC hospitalization 24640534. (\emph{Reproduces Figure~\ref{fig:mimic24640534} with detailed caption.}) This white, non-Hispanic male of age $\sim 60$ was admitted to the ED for observation. He had no previous admission history within MIMIC. His stay was 37 days 22 hours in duration. After the stay, he subsequently received ICD-10 diagnoses C9200 for `Acute myeloblastic leukemia, not having achieved remission', I214 for `Non-ST elevation (NSTEMI) myocardial infarction', D701 for `Agranulocytosis secondary to cancer chemotherapy', E222 for `Syndrome of inappropriate secretion of antidiuretic hormone', J8410 for `Pulmonary fibrosis, unspecified', K1231 for `Oral mucositis (ulcerative) due to antineoplastic therapy', T451X5A for `Adverse effect of antineoplastic and immunosuppressive drugs, initial encounter', Y92230 for `Patient room in hospital as the place of occurrence of the external cause', F4323 for `Adjustment disorder with mixed anxiety and depressed mood', R5081 for `Fever presenting with conditions classified elsewhere', F329 for `Major depressive disorder, single episode, unspecified', F952 for `Tourette's disorder', K219 for `Gastro-esophageal reflux disease without esophagitis', F17210 for `Nicotine dependence, cigarettes, uncomplicated', R740 for `Nonspecific elevation of levels of transaminase and lactic acid dehydrogenase [LDH]', R918 for `Other nonspecific abnormal finding of lung field', Z006 for `Encounter for examination for normal comparison and control in clinical research program', R2241 for `Localized swelling, mass and lump, right lower limb', Z806 for `Family history of leukemia', R197 for `Diarrhea, unspecified', I2510 for `Atherosclerotic heart disease of native coronary artery without angina pectoris', and R21 for `Rash and other nonspecific skin eruption'.}
		\label{fig:mimic24640534a}
	\end{figure}

	\newpage

	\begin{figure}[htb]
		\centering
		\includegraphics[width=0.8\textwidth]{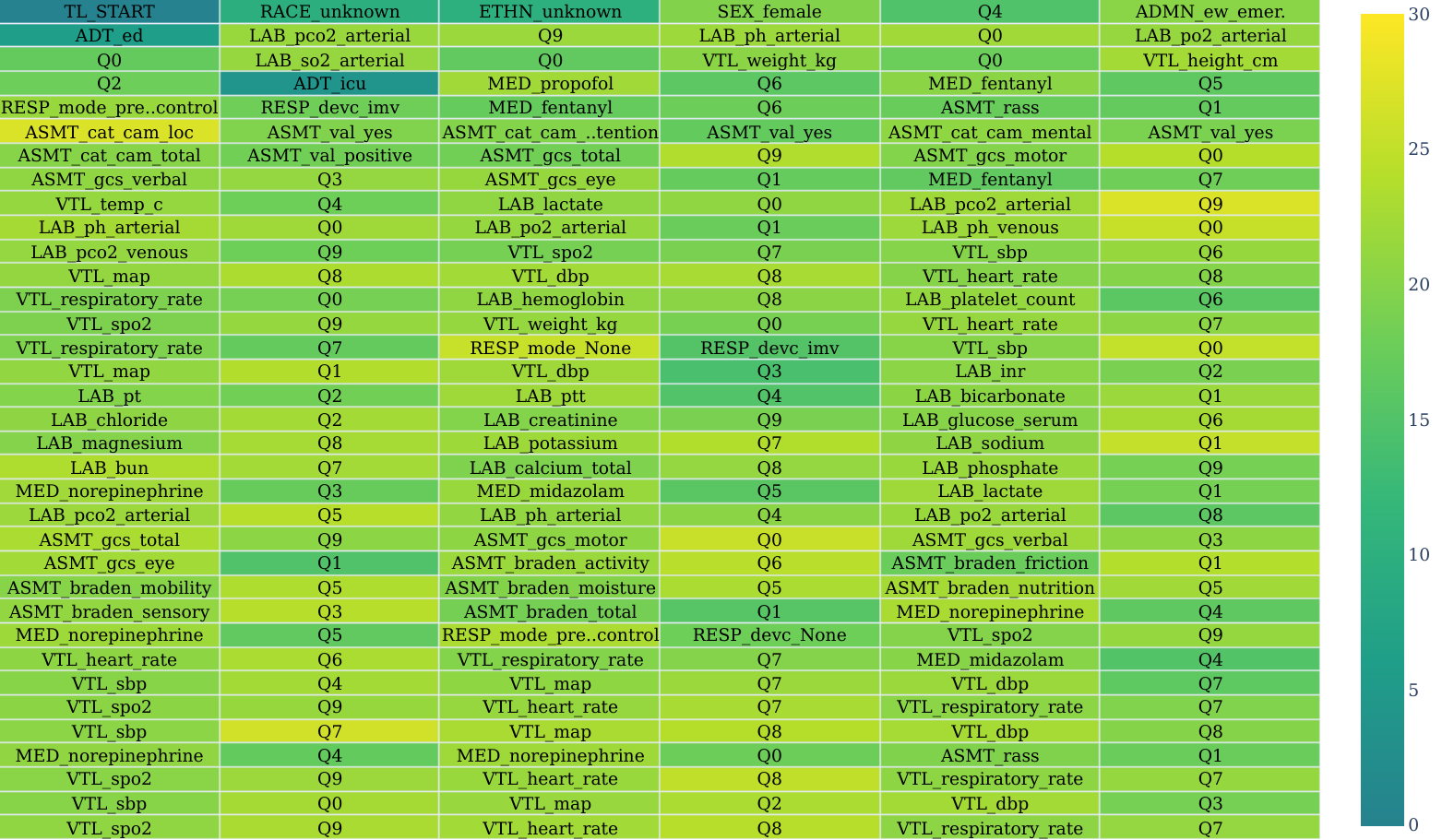}
		\caption{Tokenwise context-aware information for the first 210 tokens of MIMIC hospitalization 26886976. (\emph{Reproduces Figure~\ref{fig:mimic26886976} with detailed caption.}) This female of unknown race and ethnicity was admitted to the ED at age $\sim 73$. Her 12 day 2 hour hospital stay ended in death. After the stay, she subsequently received ICD-10 diagnoses A4189 for `Other specified sepsis', R6521 for `Severe sepsis with septic shock', J9601 for `Acute respiratory failure with hypoxia', J690 for `Pneumonitis due to inhalation of food and vomit', K810 for `Acute cholecystitis', E872 for `Acidosis', D689 for `Coagulation defect, unspecified', N179 for `Acute kidney failure, unspecified', N390 for `Urinary tract infection, site not specified', G9340 for `Encephalopathy, unspecified', E870 for `Hyperosmolality and hypernatremia', F05 for `Delirium due to known physiological condition', Z66 for `Do not resuscitate', Z515 for `Encounter for palliative care', I10 for `Essential (primary) hypertension', E119 for `Type 2 diabetes mellitus without complications', E039 for `Hypothyroidism, unspecified', K82A1 for `Gangrene of gallbladder in cholecystitis', D696 for `Thrombocytopenia, unspecified', K219 for `Gastro-esophageal reflux disease without esophagitis', F1010 for `Alcohol abuse, uncomplicated', E8339 for `Other disorders of phosphorus metabolism', E8770 for `Fluid overload, unspecified', B964 for `Proteus (mirabilis) (morganii) as the cause of diseases classified elsewhere', and K7460 for `Unspecified cirrhosis of liver'.}
		\label{fig:mimic26886976a}
	\end{figure}

    	\begin{figure}[htb]
		\centering
		\includegraphics[width=0.8\textwidth]{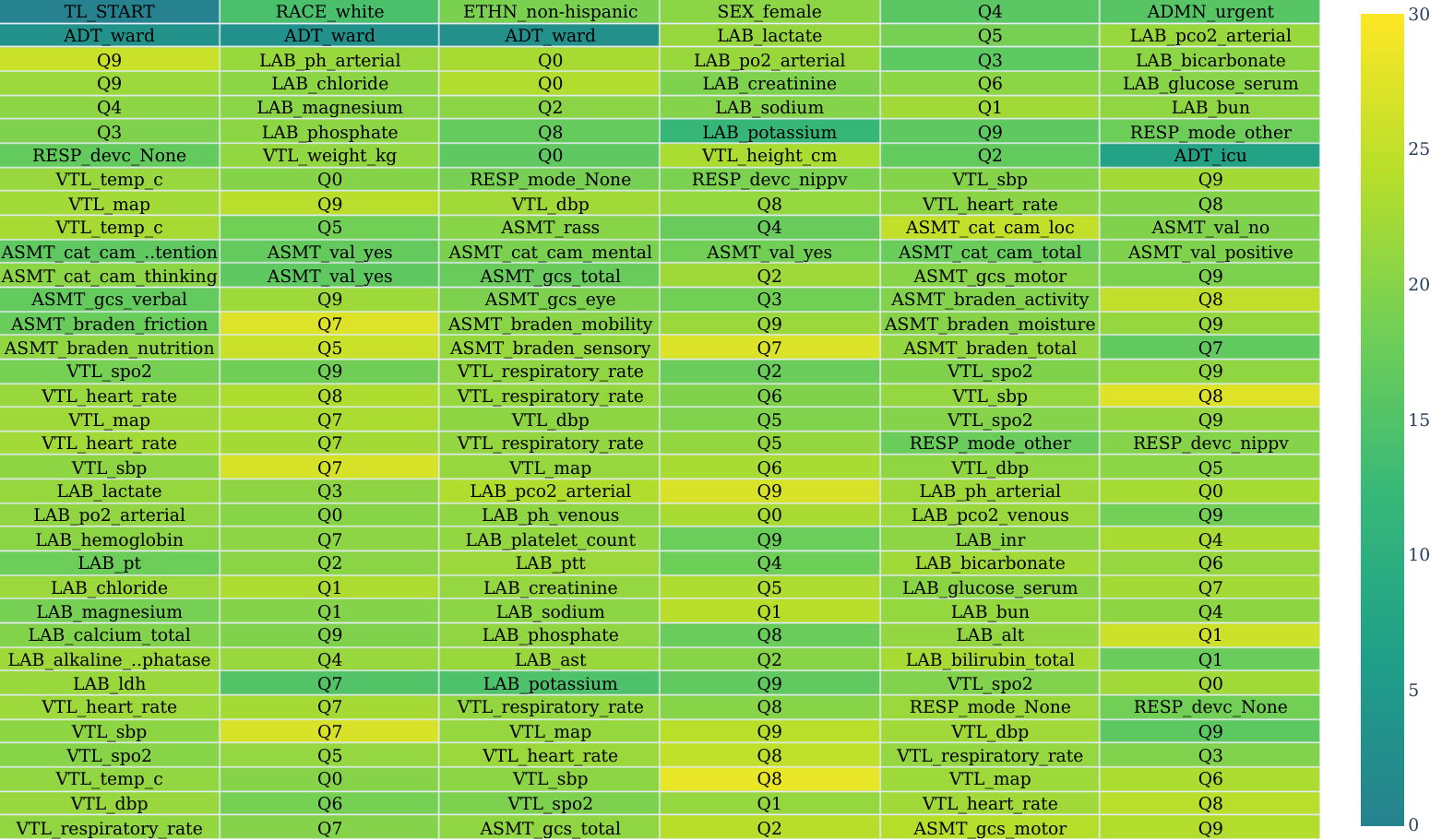}
		\caption{Tokenwise context-aware information for the first 210 tokens of MIMIC hospitalization 29022625. (\emph{Reproduces Figure~\ref{fig:mimic29022625} with detailed caption.}) This $\sim 55$ year old white female had previously been seen for  a myriad of conditions (see following page). After a a 30 day 20 hour stay, she received new diagnoses:  K2211 for `Ulcer of esophagus with bleeding',  J690 for `Pneumonitis due to inhalation of food and vomit',  J9601 for `Acute respiratory failure with hypoxia',  J9602 for `Acute respiratory failure with hypercapnia',  A419 for `Sepsis, unspecified organism',  J90 for `Pleural effusion, not elsewhere classified',  E872 for `Acidosis',  J95851 for `Ventilator associated pneumonia',  J939 for `Pneumothorax, unspecified',  E2749 for `Other adrenocortical insufficiency',  E871 for `Hypo-osmolality and hyponatremia',  I313 for `Pericardial effusion (noninflammatory)',  G9340 for `Encephalopathy, unspecified',  A0471 for `Enterocolitis due to Clostridium difficile, recurrent',  K222 for `Esophageal obstruction',  R1319 for `Other dysphagia',  I480 for `Paroxysmal atrial fibrillation',  E890 for `Postprocedural hypothyroidism',  D630 for `Anemia in neoplastic disease',  E1165 for `Type 2 diabetes mellitus with hyperglycemia',  I952 for `Hypotension due to drugs',  T4275XA for `Adverse effect of unspecified antiepileptic and sedative-hypnotic drugs, initial encounter',  K5903 for `Drug induced constipation',  T40605A for `Adverse effect of unspecified narcotics, initial encounter',  T424X5A for `Adverse effect of benzodiazepines, initial encounter',  E875 for `Hyperkalemia',  and E11649 for `Type 2 diabetes mellitus with hypoglycemia without coma'. }
		\label{fig:mimic29022625a}
	\end{figure}

	\clearpage

	\small Previous diagnoses for MIMIC hospitalization 29022625 (see Figure~\ref{fig:mimic29022625a}): K9422 for `Gastrostomy infection',  N179 for `Acute kidney failure, unspecified',  I5032 for `Chronic diastolic (congestive) heart failure',  J9612 for `Chronic respiratory failure with hypercapnia',  I82C12 for `Acute embolism and thrombosis of left internal jugular vein',  C155 for `Malignant neoplasm of lower third of esophagus',  C770 for `Secondary and unspecified malignant neoplasm of lymph nodes of head, face and neck',  Q211 for `Atrial septal defect',  R64 for `Cachexia',  L03311 for `Cellulitis of abdominal wall',  I2510 for `Atherosclerotic heart disease of native coronary artery without angina pectoris',  Z955 for `Presence of coronary angioplasty implant and graft',  E785 for `Hyperlipidemia, unspecified',  F17210 for `Nicotine dependence, cigarettes, uncomplicated',  J449 for `Chronic obstructive pulmonary disease, unspecified',  Z9981 for `Dependence on supplemental oxygen',  Z953 for `Presence of xenogenic heart valve',  F419 for `Anxiety disorder, unspecified',  F329 for `Major depressive disorder, single episode, unspecified',  M109 for `Gout, unspecified',  Z8571 for `Personal history of Hodgkin lymphoma',  E039 for `Hypothyroidism, unspecified',  I340 for `Nonrheumatic mitral (valve) insufficiency',  E1122 for `Type 2 diabetes mellitus with diabetic chronic kidney disease',  N189 for `Chronic kidney disease, unspecified',  Z794 for `Long term (current) use of insulin',  Z6820 for `Body mass index [BMI] 20.0-20.9, adult',  Y833 for `Surgical operation with formation of external stoma as the cause of abnormal reaction of the patient, or of later complication, without mention of misadventure at the time of the procedure',  Y92009 for `Unspecified place in unspecified non-institutional (private) residence as the place of occurrence of the external cause',  B372 for `Candidiasis of skin and nail',  E1151 for `Type 2 diabetes mellitus with diabetic peripheral angiopathy without gangrene',  Z7901 for `Long term (current) use of anticoagulants', C155 for `Malignant neoplasm of lower third of esophagus',  J9621 for `Acute and chronic respiratory failure with hypoxia',  J9622 for `Acute and chronic respiratory failure with hypercapnia',  E43 for `Unspecified severe protein-calorie malnutrition',  G9341 for `Metabolic encephalopathy',  I5033 for `Acute on chronic diastolic (congestive) heart failure',  I82C12 for `Acute embolism and thrombosis of left internal jugular vein',  R7881 for `Bacteremia',  J441 for `Chronic obstructive pulmonary disease with (acute) exacerbation',  Q211 for `Atrial septal defect',  J9811 for `Atelectasis',  R042 for `Hemoptysis',  J918 for `Pleural effusion in other conditions classified elsewhere',  R112 for `Nausea with vomiting, unspecified',  E860 for `Dehydration',  R1310 for `Dysphagia, unspecified',  B9689 for `Other specified bacterial agents as the cause of diseases classified elsewhere',  N189 for `Chronic kidney disease, unspecified',  Z954 for `Presence of other heart-valve replacement',  I2510 for `Atherosclerotic heart disease of native coronary artery without angina pectoris',  Z9861 for `Coronary angioplasty status',  I69998 for `Other sequelae following unspecified cerebrovascular disease',  H538 for `Other visual disturbances',  F419 for `Anxiety disorder, unspecified',  F329 for `Major depressive disorder, single episode, unspecified',  E1121 for `Type 2 diabetes mellitus with diabetic nephropathy',  I340 for `Nonrheumatic mitral (valve) insufficiency',  E1151 for `Type 2 diabetes mellitus with diabetic peripheral angiopathy without gangrene',  I739 for `Peripheral vascular disease, unspecified',  D72829 for `Elevated white blood cell count, unspecified',  D649 for `Anemia, unspecified',  E039 for `Hypothyroidism, unspecified',  Z03818 for `Encounter for observation for suspected exposure to other biological agents ruled out',  Z6821 for `Body mass index [BMI] 21.0-21.9, adult',  E785 for `Hyperlipidemia, unspecified',  D563 for `Thalassemia minor',  M1A9XX0 for `Chronic gout, unspecified, without tophus (tophi)',  Z720 for `Tobacco use',  and T451X5A for `Adverse effect of antineoplastic and immunosuppressive drugs, initial encounter'.

	\begin{figure}[htb]
		\centering
		\includegraphics[width=0.8\textwidth]{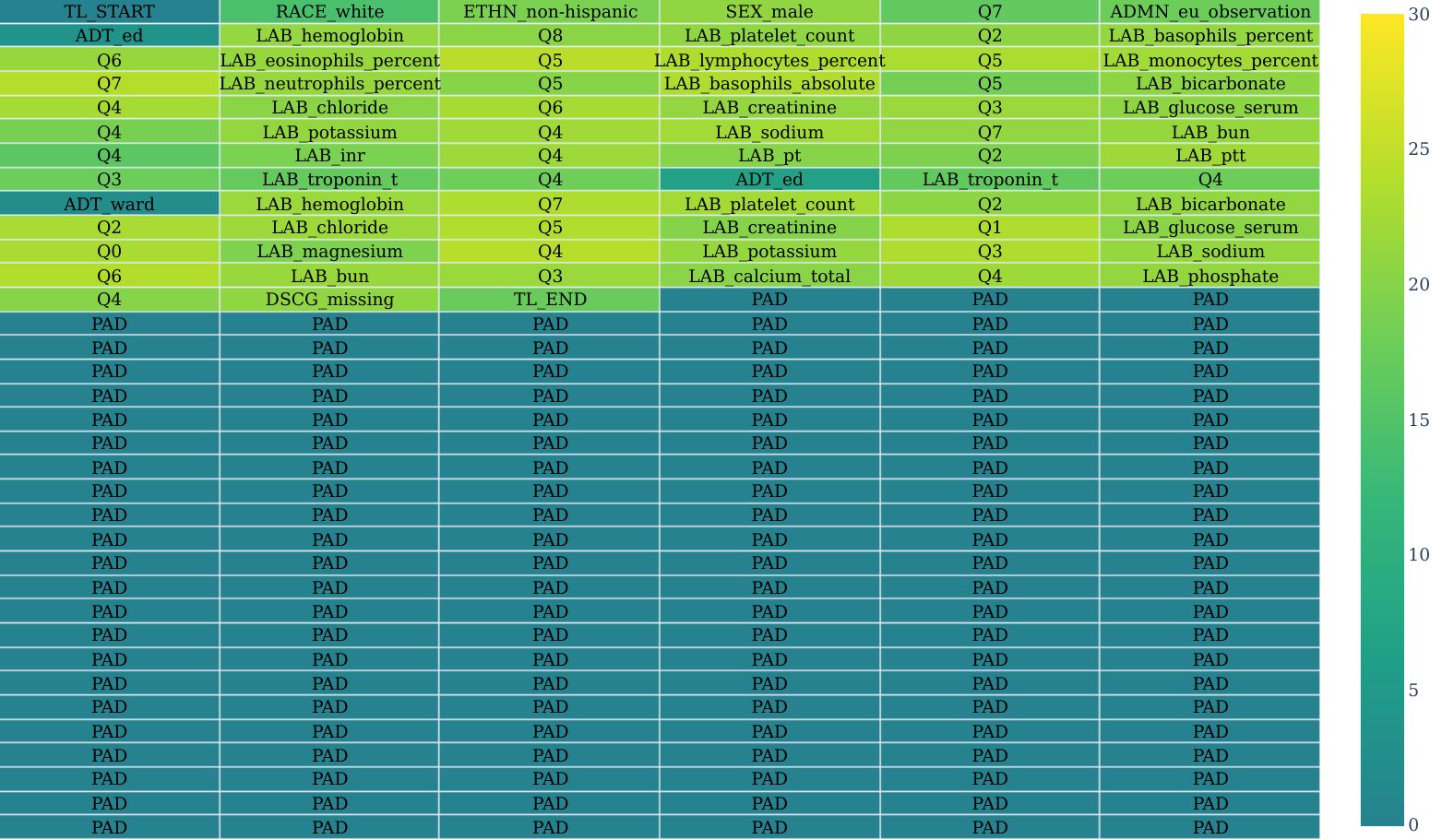}
		\caption{Tokenwise context-aware information for the first 210 tokens of MIMIC hospitalization 29173149. This $\sim 75$ year old white male had previously been seen for K388 for `Other specified diseases of appendix', M4854XA for `Collapsed vertebra, not elsewhere classified, thoracic region, initial encounter for fracture', G20 for `Parkinson's disease', G250 for `Essential tremor', R2681 for `Unsteadiness on feet', Z8547 for `Personal history of malignant neoplasm of testis', Z87891 for `Personal history of nicotine dependence', and F329 for `Major depressive disorder, single episode, unspecified'. After a 1 day 14 hour long admission, he received diagnoses: S0003XA for `Contusion of scalp, initial encounter', W1839XA for `Other fall on same level, initial encounter', Y92091 for `Bathroom in other non-institutional residence as the place of occurrence of the external cause', Z9181 for `History of falling', E785 for `Hyperlipidemia, unspecified', F419 for `Anxiety disorder, unspecified', Z820 for `Family history of epilepsy and other diseases of the nervous system', R319 for `Hematuria, unspecified', and D696 for `Thrombocytopenia, unspecified'. }
		\label{fig:mimic29173149}
	\end{figure}

	\begin{figure}[htb]
		\centering
		\includegraphics[width=0.8\textwidth]{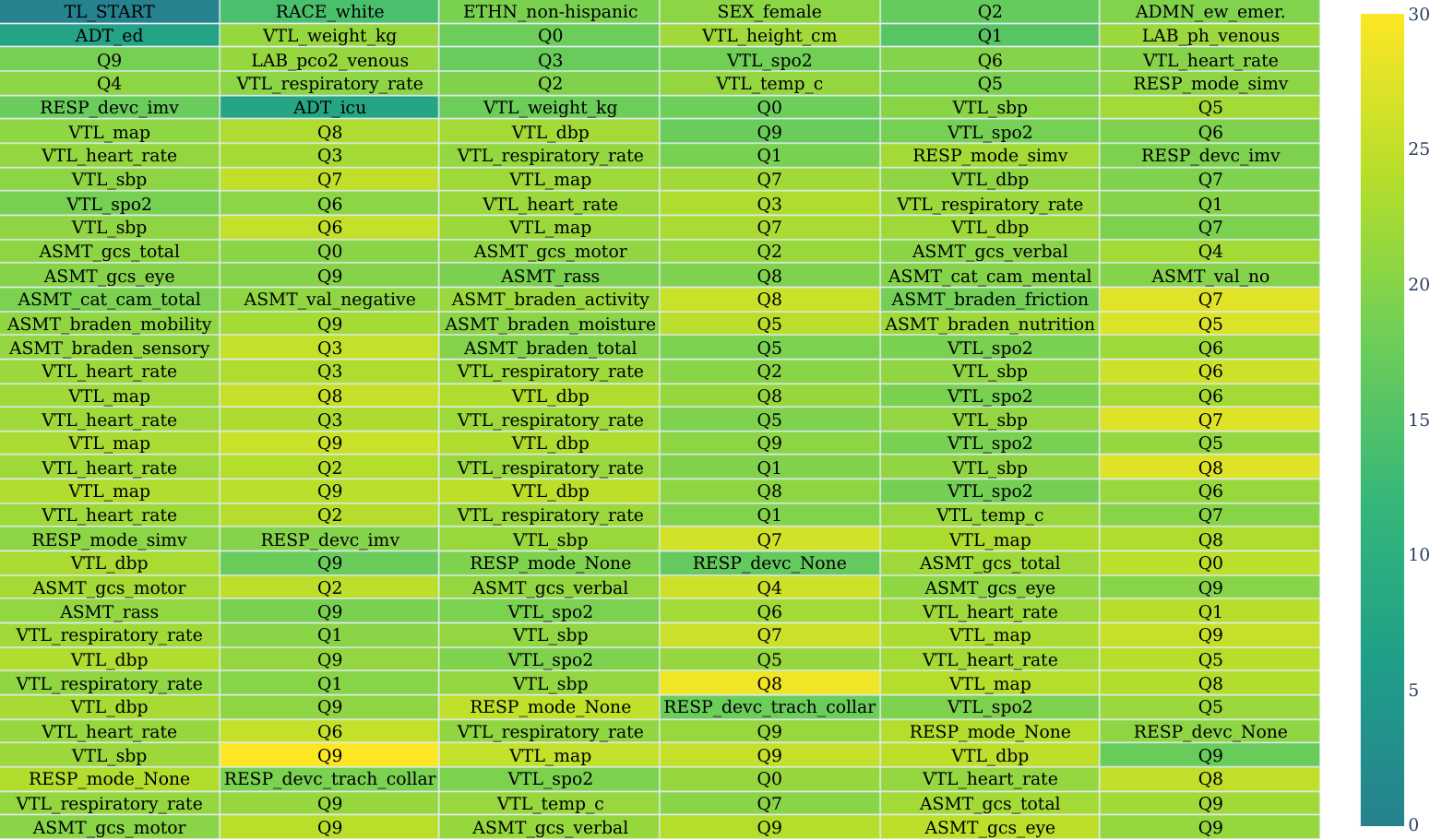}
		\caption{Tokenwise context-aware information for the first 210 tokens of MIMIC hospitalization 27267707. This $\sim 41$ year old female had previously been seen for: N200 for `Calculus of kidney',  G92 for `Toxic encephalopathy',  J9610 for `Chronic respiratory failure, unspecified whether with hypoxia or hypercapnia',  G114 for `Hereditary spastic paraplegia',  Z930 for `Tracheostomy status',  N1330 for `Unspecified hydronephrosis',  N12 for `Tubulo-interstitial nephritis, not specified as acute or chronic',  N3090 for `Cystitis, unspecified without hematuria',  Q639 for `Congenital malformation of kidney, unspecified',  G809 for `Cerebral palsy, unspecified',  E233 for `Hypothalamic dysfunction, not elsewhere classified',  R110 for `Nausea',  Z993 for `Dependence on wheelchair',  Z981 for `Arthrodesis status',  N200 for `Calculus of kidney',  G800 for `Spastic quadriplegic cerebral palsy',  Z9911 for `Dependence on respirator [ventilator] status',  J9610 for `Chronic respiratory failure, unspecified whether with hypoxia or hypercapnia',  N1330 for `Unspecified hydronephrosis',  N390 for `Urinary tract infection, site not specified',  Z930 for `Tracheostomy status',  Q639 for `Congenital malformation of kidney, unspecified',  Z993 for `Dependence on wheelchair',  Z981 for `Arthrodesis status',  R32 for `Unspecified urinary incontinence',  5920 for `Calculus of kidney',  51883 for `Chronic respiratory failure',  V440 for `Tracheostomy status',  2762 for `Acidosis',  591 for `Hydronephrosis',  3341 for `Hereditary spastic paraplegia',  V463 for `Wheelchair dependence',  28860 for `Leukocytosis, unspecified',  49390 for `Asthma, unspecified type, unspecified',  3159 for `Unspecified delay in development',  V454 for `Arthrodesis status',  6268 for `Other disorders of menstruation and other abnormal bleeding from female genital tract',  3432 for `Congenital quadriplegia',  V463 for `Wheelchair dependence',  V550 for `Attention to tracheostomy',  49390 for `Asthma, unspecified type, unspecified',  51889 for `Other diseases of lung, not elsewhere classified',  V4611 for `Dependence on respirator, status',  7533 for `Other specified anomalies of kidney',  30981 for `Posttraumatic stress disorder',  5859 for `Chronic kidney disease, unspecified',  and V074 for `Hormone replacement therapy (postmenopausal)'. After a 2 day 19 hour stay, she subsequently received new diagnoses: J208 for `Acute bronchitis due to other specified organisms',  J9621 for `Acute and chronic respiratory failure with hypoxia',  E872 for `Acidosis',  Z9981 for `Dependence on supplemental oxygen',  E2839 for `Other primary ovarian failure',  Z905 for `Acquired absence of kidney',  J45909 for `Unspecified asthma, uncomplicated',  M419 for `Scoliosis, unspecified',  M810 for `Age-related osteoporosis without current pathological fracture',  F4310 for `Post-traumatic stress disorder, unspecified',  F39 for `Unspecified mood [affective] disorder',  Q632 for `Ectopic kidney',  N189 for `Chronic kidney disease, unspecified',  I517 for `Cardiomegaly'.}
		\label{fig:mimic27267707}
	\end{figure}

	\clearpage
	\appendix{Example timelines from UCMC}

	\begin{figure}[htb]
		\centering
		\includegraphics[width=0.8\textwidth]{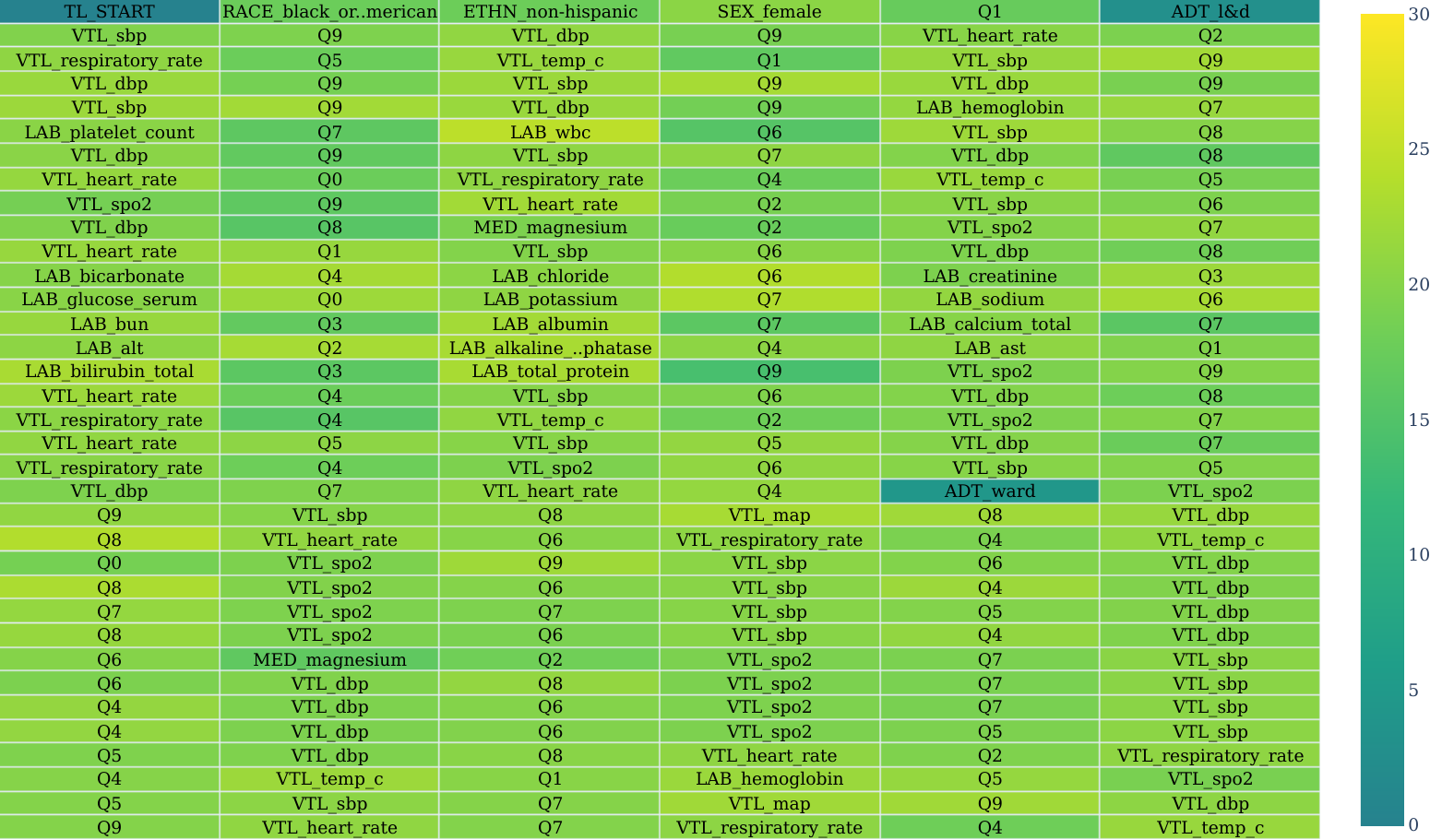}
		\caption{Tokenwise context-aware information for the first 210 tokens of UCMC hospitalization 8797520.}
		\label{fig:mimic8797520}
	\end{figure}

	\begin{figure}[htb]
		\centering
		\includegraphics[width=0.8\textwidth]{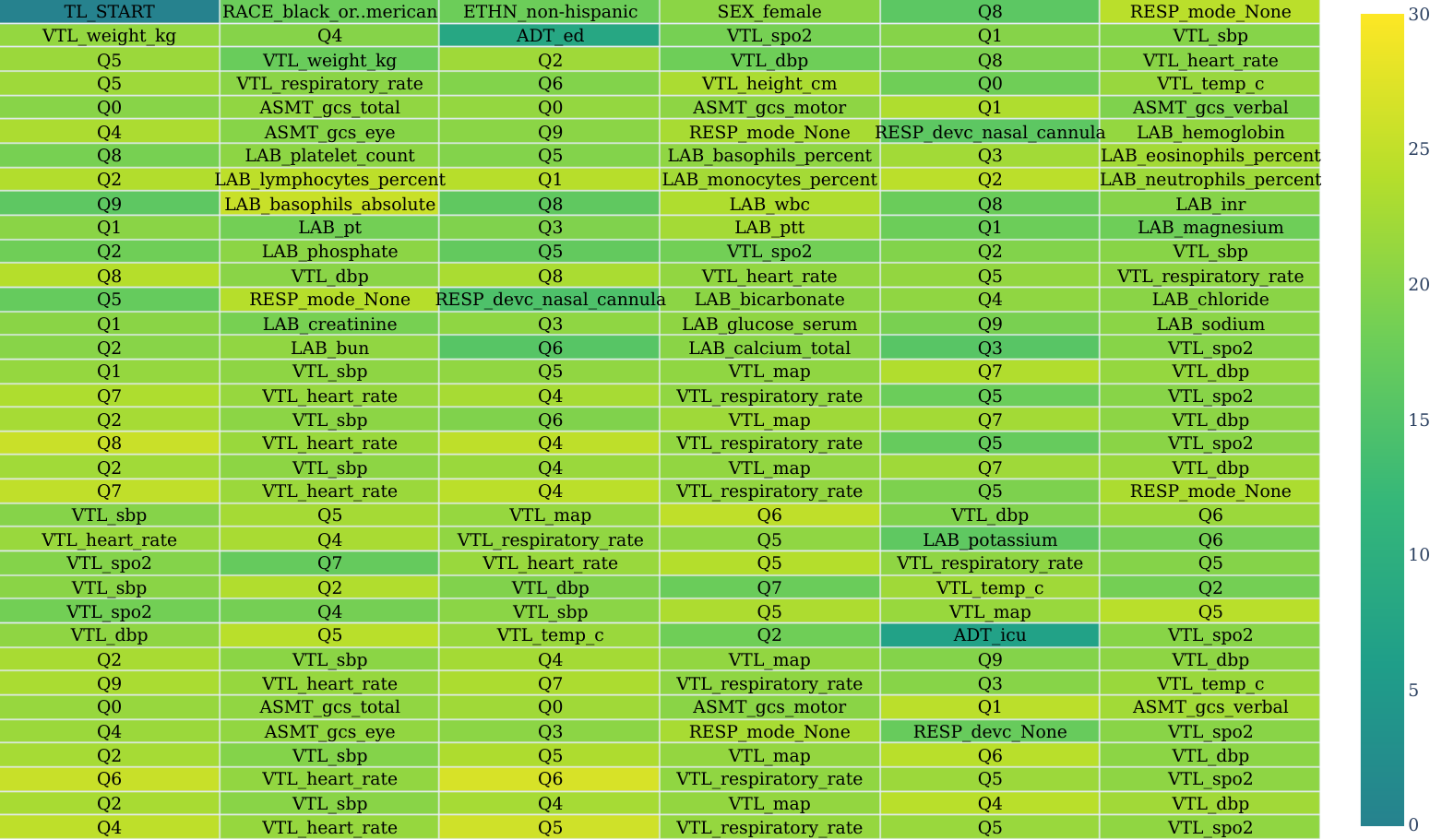}
		\caption{Tokenwise context-aware information for the first 210 tokens of UCMC hospitalization 27055120.}
		\label{fig:mimic27055120}
	\end{figure}

	\begin{figure}[htb]
		\centering
		\includegraphics[width=0.8\textwidth]{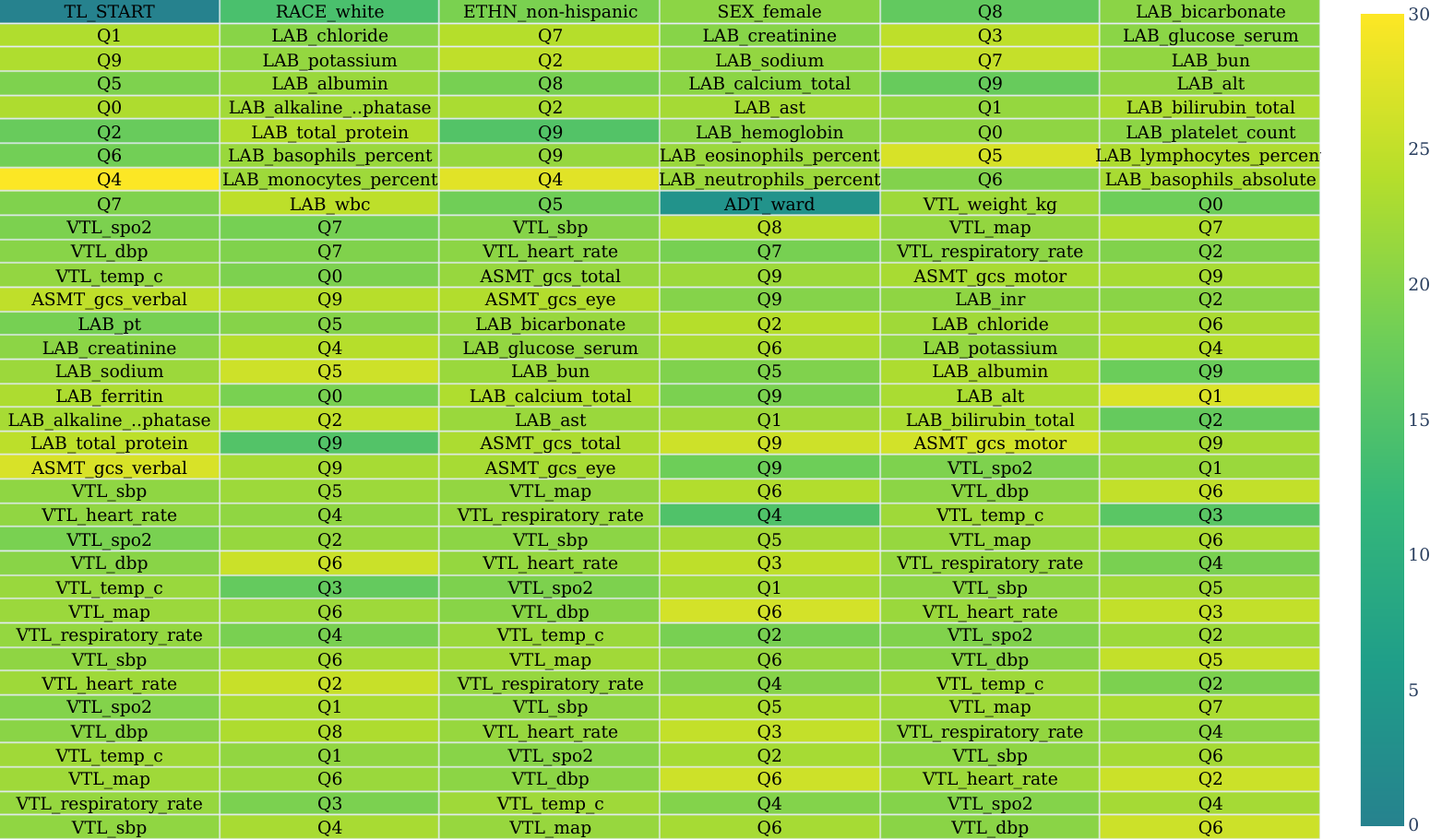}
		\caption{Tokenwise context-aware information for the first 210 tokens of UCMC hospitalization 10969205.}
		\label{fig:mimic10969205}
	\end{figure}

	\begin{figure}[htb]
		\centering
		\includegraphics[width=0.8\textwidth]{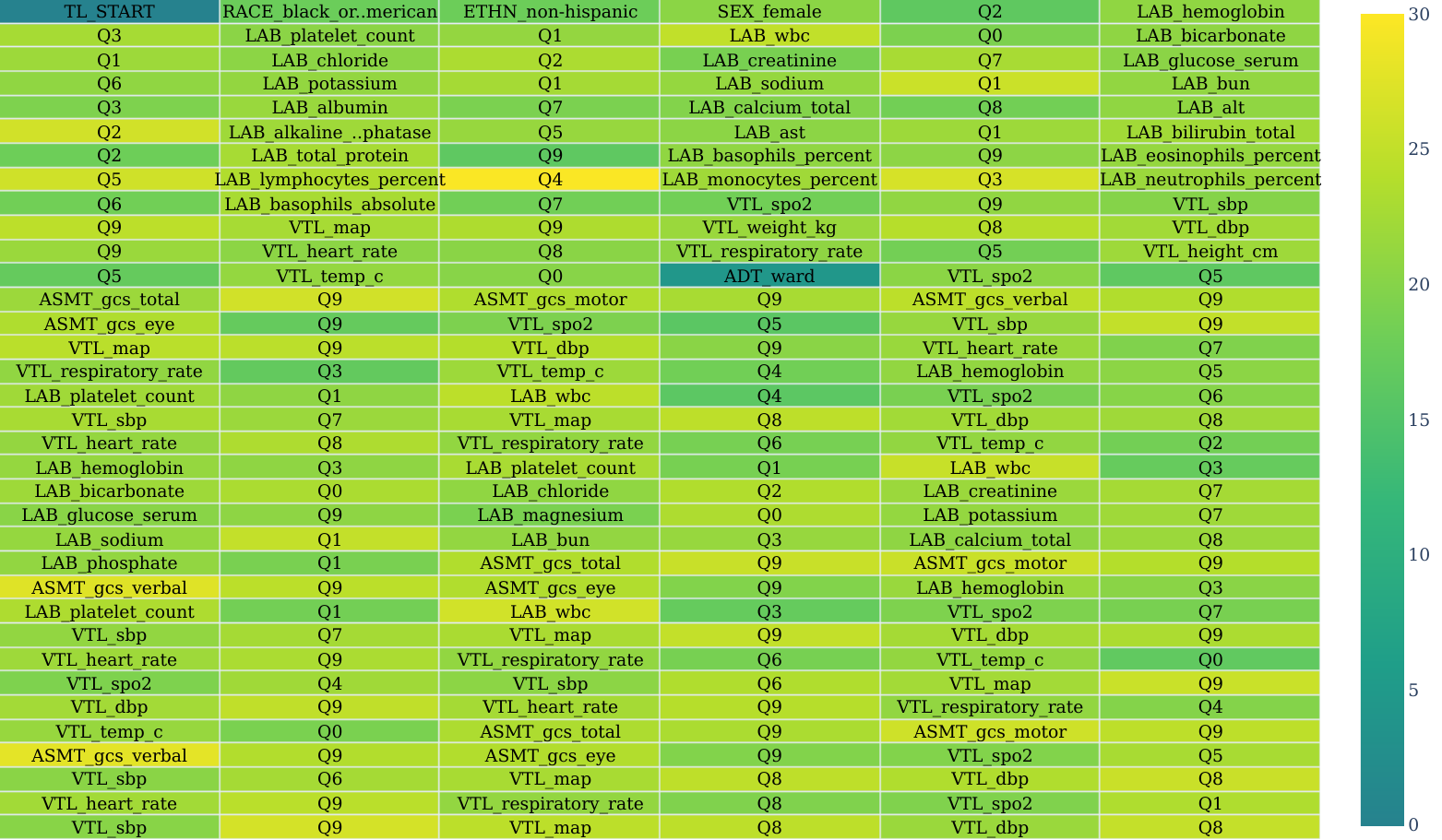}
		\caption{Tokenwise context-aware information for the first 210 tokens of UCMC hospitalization 2974992.}
		\label{fig:mimic2974992}
	\end{figure}

	\begin{figure}[htb]
		\centering
		\includegraphics[width=0.8\textwidth]{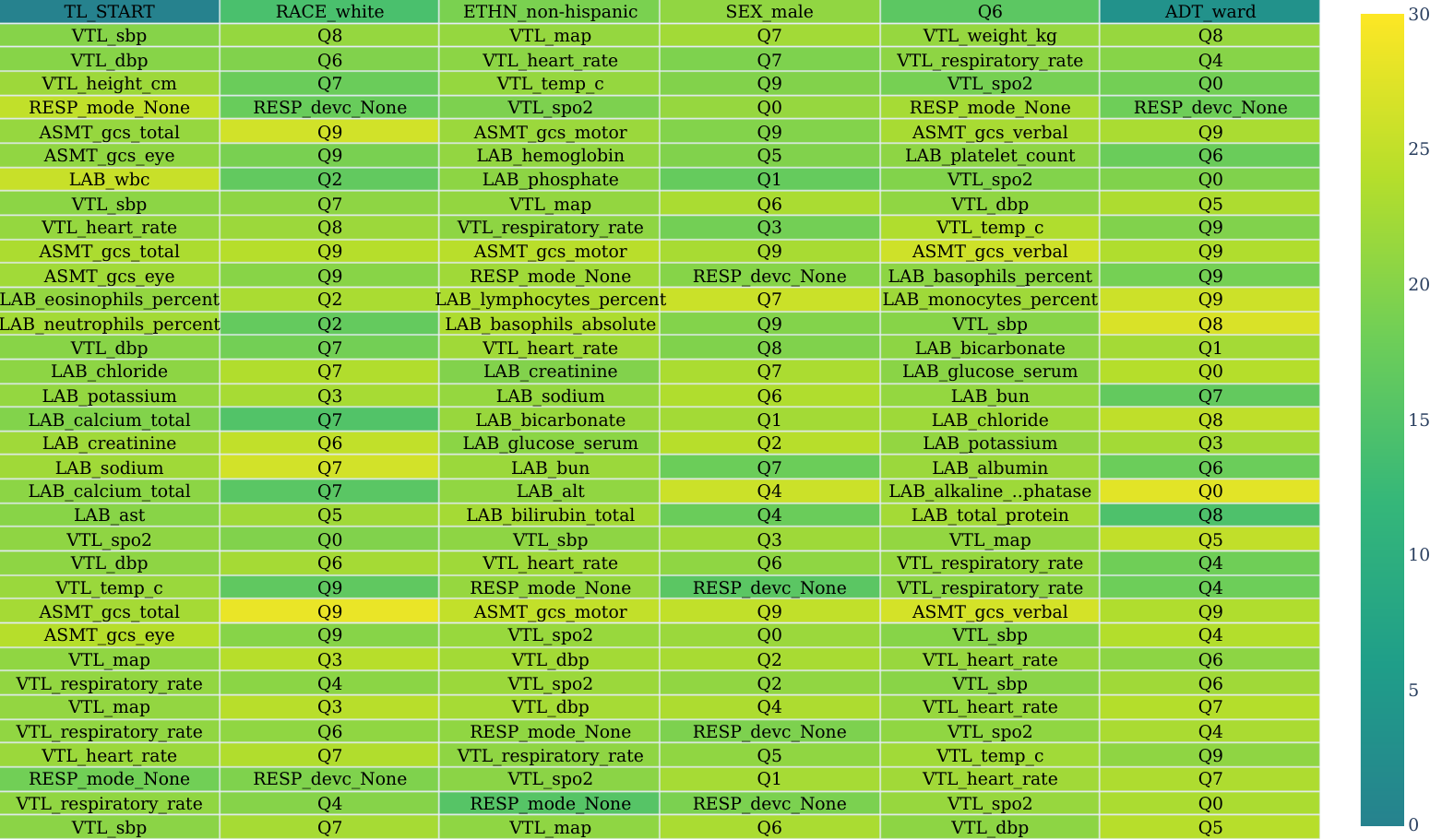}
		\caption{Tokenwise context-aware information for the first 210 tokens of UCMC hospitalization 20528107.}
		\label{fig:mimic20528107}
	\end{figure}

	\clearpage
	\appendix{Full results for the redaction experiments}
    \label{s:full_results}
    This section contains all tables of results discussed in \S\ref{ss:redaction_results}.

	\begin{table}[htb]
		\tbl{Performance metrics for inpatient mortality prediction task on ICU patients in the MIMIC test set. The $p$-val. columns indicate the results of a hypothesis test for the corresponding metric against the one-sided alternative that the model trained on the original data performs better. Blank entries indicate $p>0.1$.}
		{
			\begin{tabular}{lllrlrlr}
				\toprule
				\multicolumn{2}{c}{version} & \multicolumn{2}{c}{ROC-AUC} & \multicolumn{2}{c}{PR-AUC} & \multicolumn{2}{c}{Brier}                                                               \\
				\cmidrule(lr){1-2} \cmidrule(lr){3-4} \cmidrule(lr){5-6} \cmidrule(lr){7-8}
				method                      & pct.                        & range                      & $p$-val.                  & range             & $p$-val. & range             & $p$-val. \\ \midrule
				original                    & ---                         & $0.869 \pm 0.009$          & ---                       & $0.464 \pm 0.032$ & ---      & $0.065 \pm 0.003$ & ---      \\ \midrule
				\multirow{4}{*}{top}        & 10                          & $0.860 \pm 0.010$          &                           & $0.440 \pm 0.028$ & 0.088    & $0.067 \pm 0.003$ &          \\
				                            & 20                          & $0.848 \pm 0.011$          & 0.003                     & $0.429 \pm 0.031$ & 0.029    & $0.068 \pm 0.003$ &          \\
				                            & 30                          & $0.833 \pm 0.012$          & $<0.001$                  & $0.395 \pm 0.034$ & 0.001    & $0.069 \pm 0.004$ & 0.031    \\
				                            & 40                          & $0.823 \pm 0.011$          & $<0.001$                  & $0.386 \pm 0.031$ & $<0.001$ & $0.070 \pm 0.004$ & 0.008    \\ \midrule
				\multirow{4}{*}{bottom}     & 10                          & $0.867 \pm 0.010$          &                           & $0.465 \pm 0.025$ &          & $0.065 \pm 0.003$ &          \\
				                            & 20                          & $0.866 \pm 0.009$          &                           & $0.456 \pm 0.031$ &          & $0.065 \pm 0.003$ &          \\
				                            & 30                          & $0.862 \pm 0.011$          &                           & $0.445 \pm 0.029$ &          & $0.065 \pm 0.004$ &          \\
				                            & 40                          & $0.859 \pm 0.012$          &                           & $0.451 \pm 0.030$ &          & $0.065 \pm 0.004$ &          \\ \midrule
				\multirow{4}{*}{random}     & 10                          & $0.866 \pm 0.008$          &                           & $0.456 \pm 0.026$ &          & $0.066 \pm 0.002$ &          \\
				                            & 20                          & $0.863 \pm 0.008$          &                           & $0.454 \pm 0.030$ &          & $0.066 \pm 0.003$ &          \\
				                            & 30                          & $0.865 \pm 0.010$          &                           & $0.456 \pm 0.025$ &          & $0.066 \pm 0.004$ &          \\
				                            & 40                          & $0.861 \pm 0.011$          &                           & $0.457 \pm 0.025$ &          & $0.065 \pm 0.003$ &          \\
				\bottomrule
			\end{tabular}
		}
		\label{tbl:sad_mimic}
	\end{table}

	\begin{table}[htb]
		\tbl{Performance metrics for long length of stay prediction task for ICU patients in the MIMIC test set. The $p$-val columns are as in Table~\ref{tbl:sad_mimic}.}
		{
			\begin{tabular}{lllrlrlr}
				\toprule
				\multicolumn{2}{c}{version} & \multicolumn{2}{c}{ROC-AUC} & \multicolumn{2}{c}{PR-AUC} & \multicolumn{2}{c}{Brier}                                                             \\
				\cmidrule(lr){1-2} \cmidrule(lr){3-4} \cmidrule(lr){5-6} \cmidrule(lr){7-8}
				method                      & pct.                        & range                      & p.-val                    & range             & p.-val & range             & p.-val   \\ \midrule
				original                    & ---                         & $0.740 \pm 0.008$          & ---                       & $0.657 \pm 0.015$ & ---    & $0.204 \pm 0.003$ & ---      \\ \midrule
				\multirow{4}{*}{top}        & 10                          & $0.735 \pm 0.009$          &                           & $0.654 \pm 0.016$ &        & $0.206 \pm 0.003$ &          \\
				                            & 20                          & $0.726 \pm 0.009$          & 0.009                     & $0.640 \pm 0.014$ & 0.061  & $0.208 \pm 0.004$ & 0.019    \\
				                            & 30                          & $0.720 \pm 0.009$          & 0.001                     & $0.633 \pm 0.012$ & 0.013  & $0.211 \pm 0.004$ & 0.001    \\
				                            & 40                          & $0.714 \pm 0.009$          & $<0.001$                  & $0.638 \pm 0.013$ & 0.017  & $0.212 \pm 0.003$ & $<0.001$ \\ \midrule
				\multirow{4}{*}{bottom}     & 10                          & $0.736 \pm 0.012$          &                           & $0.652 \pm 0.015$ &        & $0.206 \pm 0.005$ &          \\
				                            & 20                          & $0.732 \pm 0.010$          &                           & $0.650 \pm 0.016$ &        & $0.207 \pm 0.004$ &          \\
				                            & 30                          & $0.726 \pm 0.009$          & 0.018                     & $0.646 \pm 0.014$ &        & $0.209 \pm 0.004$ & 0.032    \\
				                            & 40                          & $0.724 \pm 0.008$          & 0.005                     & $0.644 \pm 0.013$ &        & $0.209 \pm 0.003$ & 0.013    \\ \midrule
				\multirow{4}{*}{random}     & 10                          & $0.737 \pm 0.006$          &                           & $0.654 \pm 0.013$ &        & $0.205 \pm 0.003$ &          \\
				                            & 20                          & $0.733 \pm 0.009$          & 0.099                     & $0.647 \pm 0.014$ &        & $0.206 \pm 0.004$ &          \\
				                            & 30                          & $0.728 \pm 0.007$          & 0.053                     & $0.647 \pm 0.011$ &        & $0.208 \pm 0.002$ & 0.076    \\
				                            & 40                          & $0.727 \pm 0.008$          & 0.020                     & $0.645 \pm 0.014$ & 0.089  & $0.208 \pm 0.003$ & 0.034    \\
				\bottomrule
			\end{tabular}
		}
		\label{tbl:llos_mimic}
	\end{table}

	\begin{table}[htb]
		\tbl{Performance metrics for inpatient mortality prediction task for ICU patients in the UCMC test dataset. The $p$-val columns are as in Table~\ref{tbl:sad_mimic}.}
		{
			\begin{tabular}{lllrlrlr}
				\toprule
				\multicolumn{2}{c}{version} & \multicolumn{2}{c}{ROC-AUC} & \multicolumn{2}{c}{PR-AUC} & \multicolumn{2}{c}{Brier}                                                           \\
				\cmidrule(lr){1-2} \cmidrule(lr){3-4} \cmidrule(lr){5-6} \cmidrule(lr){7-8}
				method                      & pct.                        & range                      & p.-val                    & range             & p.-val & range             & p.-val \\ \midrule
				original                    & ---                         & $0.839 \pm 0.013$          & ---                       & $0.425 \pm 0.036$ & ---    & $0.087 \pm 0.005$ & ---    \\ \midrule
				\multirow{4}{*}{top}        & 10                          & $0.830 \pm 0.013$          &                           & $0.432 \pm 0.026$ &        & $0.088 \pm 0.004$ &        \\
				                            & 20                          & $0.814 \pm 0.014$          & 0.006                     & $0.403 \pm 0.029$ &        & $0.090 \pm 0.004$ &        \\
				                            & 30                          & $0.812 \pm 0.013$          & 0.004                     & $0.395 \pm 0.032$ &        & $0.093 \pm 0.004$ & 0.016  \\
				                            & 40                          & $0.818 \pm 0.012$          & 0.018                     & $0.395 \pm 0.029$ & 0.076  & $0.094 \pm 0.004$ & 0.006  \\ \midrule
				\multirow{4}{*}{bottom}     & 10                          & $0.834 \pm 0.011$          &                           & $0.416 \pm 0.029$ &        & $0.088 \pm 0.004$ &        \\
				                            & 20                          & $0.834 \pm 0.012$          &                           & $0.414 \pm 0.033$ &        & $0.087 \pm 0.005$ &        \\
				                            & 30                          & $0.829 \pm 0.012$          &                           & $0.405 \pm 0.033$ &        & $0.087 \pm 0.005$ &        \\
				                            & 40                          & $0.829 \pm 0.011$          &                           & $0.411 \pm 0.031$ &        & $0.085 \pm 0.005$ &        \\ \midrule
				\multirow{4}{*}{random}     & 10                          & $0.838 \pm 0.012$          &                           & $0.419 \pm 0.031$ &        & $0.086 \pm 0.005$ &        \\
				                            & 20                          & $0.835 \pm 0.011$          &                           & $0.413 \pm 0.033$ &        & $0.087 \pm 0.004$ &        \\
				                            & 30                          & $0.835 \pm 0.014$          &                           & $0.434 \pm 0.038$ &        & $0.085 \pm 0.005$ &        \\
				                            & 40                          & $0.835 \pm 0.011$          &                           & $0.426 \pm 0.028$ &        & $0.084 \pm 0.004$ &        \\
				\bottomrule
			\end{tabular}
		}
		\label{tbl:sad_ucmc}
	\end{table}

	\begin{table}[htb]
		\tbl{Performance metrics for long length of stay prediction task for ICU patients in the UCMC test set. The $p$-val columns are as in Table~\ref{tbl:sad_mimic}.}
		{
			\begin{tabular}{lllrlrlr}
				\toprule
				\multicolumn{2}{c}{version} & \multicolumn{2}{c}{ROC-AUC} & \multicolumn{2}{c}{PR-AUC} & \multicolumn{2}{c}{Brier}                                                           \\
				\cmidrule(lr){1-2} \cmidrule(lr){3-4} \cmidrule(lr){5-6} \cmidrule(lr){7-8}
				method                      & pct.                        & range                      & p.-val                    & range             & p.-val & range             & p.-val \\ \midrule
				original                    & ---                         & $0.661 \pm 0.011$          & ---                       & $0.639 \pm 0.016$ & ---    & $0.235 \pm 0.004$ & ---    \\ \midrule
				\multirow{4}{*}{top}        & 10                          & $0.671 \pm 0.011$          &                           & $0.657 \pm 0.017$ &        & $0.233 \pm 0.004$ &        \\
				                            & 20                          & $0.653 \pm 0.012$          &                           & $0.635 \pm 0.016$ &        & $0.237 \pm 0.004$ &        \\
				                            & 30                          & $0.642 \pm 0.011$          & 0.028                     & $0.631 \pm 0.015$ &        & $0.239 \pm 0.004$ &        \\
				                            & 40                          & $0.649 \pm 0.012$          & 0.094                     & $0.638 \pm 0.017$ &        & $0.238 \pm 0.004$ & 0.070  \\ \midrule
				\multirow{4}{*}{bottom}     & 10                          & $0.659 \pm 0.013$          &                           & $0.639 \pm 0.016$ &        & $0.235 \pm 0.005$ &        \\
				                            & 20                          & $0.667 \pm 0.012$          &                           & $0.647 \pm 0.019$ &        & $0.231 \pm 0.004$ &        \\
				                            & 30                          & $0.667 \pm 0.013$          &                           & $0.646 \pm 0.016$ &        & $0.231 \pm 0.004$ &        \\
				                            & 40                          & $0.664 \pm 0.011$          &                           & $0.643 \pm 0.018$ &        & $0.233 \pm 0.004$ &        \\ \midrule
				\multirow{4}{*}{random}     & 10                          & $0.664 \pm 0.012$          &                           & $0.644 \pm 0.016$ &        & $0.233 \pm 0.004$ &        \\
				                            & 20                          & $0.667 \pm 0.012$          &                           & $0.650 \pm 0.016$ &        & $0.231 \pm 0.004$ &        \\
				                            & 30                          & $0.674 \pm 0.009$          &                           & $0.654 \pm 0.015$ &        & $0.230 \pm 0.003$ &        \\
				                            & 40                          & $0.674 \pm 0.011$          &                           & $0.654 \pm 0.017$ &        & $0.230 \pm 0.004$ &        \\
				\bottomrule
			\end{tabular}
		}
		\label{tbl:llos_ucmc}
	\end{table}

\fi

\end{document}